\documentclass{article}

\usepackage{hyperref}

\usepackage{CJK} 

\usepackage{float} 

\usepackage{microtype}

\usepackage{graphicx}
\usepackage[skip=0pt]{caption} 
\usepackage{subcaption} 

\usepackage{booktabs} 
\usepackage{array}    
\usepackage{tabularx} 

\usepackage{enumitem} 

\usepackage[accepted]{icml2025}

\newcommand{\arxivnote}{\textbf{This is a preprint version submitted to arXiv.}}

\usepackage{amsmath}
\usepackage{amssymb}
\usepackage{mathtools}
\usepackage{amsthm}

\usepackage[capitalize,noabbrev]{cleveref}

\theoremstyle{plain}

\theoremstyle{definition}

\theoremstyle{remark}

\usepackage[textsize=tiny]{todonotes}

\icmltitlerunning{Control LLM: Controlled Evolution for Intelligence Retention in LLM  }

\begin{document}

\twocolumn[
\icmltitle{Control LLM: Controlled Evolution for Intelligence Retention in LLM }

\begin{center}
\arxivnote
\end{center}







\begin{icmlauthorlist}
\icmlauthor{Haichao Wei}{linkedin} 
\icmlauthor{Yunxiang Ren}{linkedin} 
\icmlauthor{Zhoutong Fu}{linkedin} 
\icmlauthor{Aman Lunia}{linkedin} 
\icmlauthor{Yi-Lin Chen}{linkedin} 
\icmlauthor{Alice Leung}{linkedin} 
\icmlauthor{Ya Xu}{google} 
\end{icmlauthorlist}

\icmlaffiliation{linkedin}{LinkedIn, Sunnyvale, CA, US} 
\icmlaffiliation{google}{Google, Mountain View, CA, US} 

\icmlcorrespondingauthor{Haichao Wei}{hawei@linkedin.com}

\icmlkeywords{Deep Learning, Large Language Model, Catastrophic Forgetting, Post-training, Continuous pre-training, Continuous Supervised Fine-tuning, ICML}

\vskip 0.3in
]

\begin{abstract}
Large Language Models (LLMs) demand significant computational resources, making it essential to enhance their capabilities without retraining from scratch. A key challenge in this domain is \textit{catastrophic forgetting} (CF), which hampers performance during Continuous Pre-training (CPT) and Continuous Supervised Fine-Tuning (CSFT). We propose \textbf{Control LLM}, a novel approach that leverages parallel pre-trained and expanded transformer blocks, aligning their hidden-states through interpolation strategies This method effectively preserves performance on existing tasks while seamlessly integrating new knowledge.

Extensive experiments demonstrate the effectiveness of Control LLM in both CPT and CSFT. On Llama3.1-8B-Instruct, it achieves significant improvements in mathematical reasoning ($+14.4\%$ on Math-Hard) and coding performance ($+10\%$ on MBPP-PLUS). On Llama3.1-8B, it enhances multilingual capabilities ($+10.6\%$ on C-Eval, $+6.8\%$ on CMMLU, and $+30.2\%$ on CMMLU-0shot-CoT). It surpasses existing methods and achieves SOTA among open-source models tuned from the same base model, using substantially less data and compute. Crucially, these gains are realized while preserving strong original capabilities, with minimal degradation ($<4.3\% \text{on MMLU}$) compared to $>35\%$ in open-source Math and Coding models. Control LLM has been successfully deployed in LinkedIn's GenAI-powered job seeker and Ads unit products.

To support further research, we release the training and evaluation code (\url{https://github.com/linkedin/ControlLLM}) along with models trained on public datasets (\url{ https://huggingface.co/ControlLLM}) to the community.

\end{abstract}

\section{Introduction}
\label{sec:intro}

\begin{figure}[htb]
  \centering
  \includegraphics[width=0.95\linewidth]{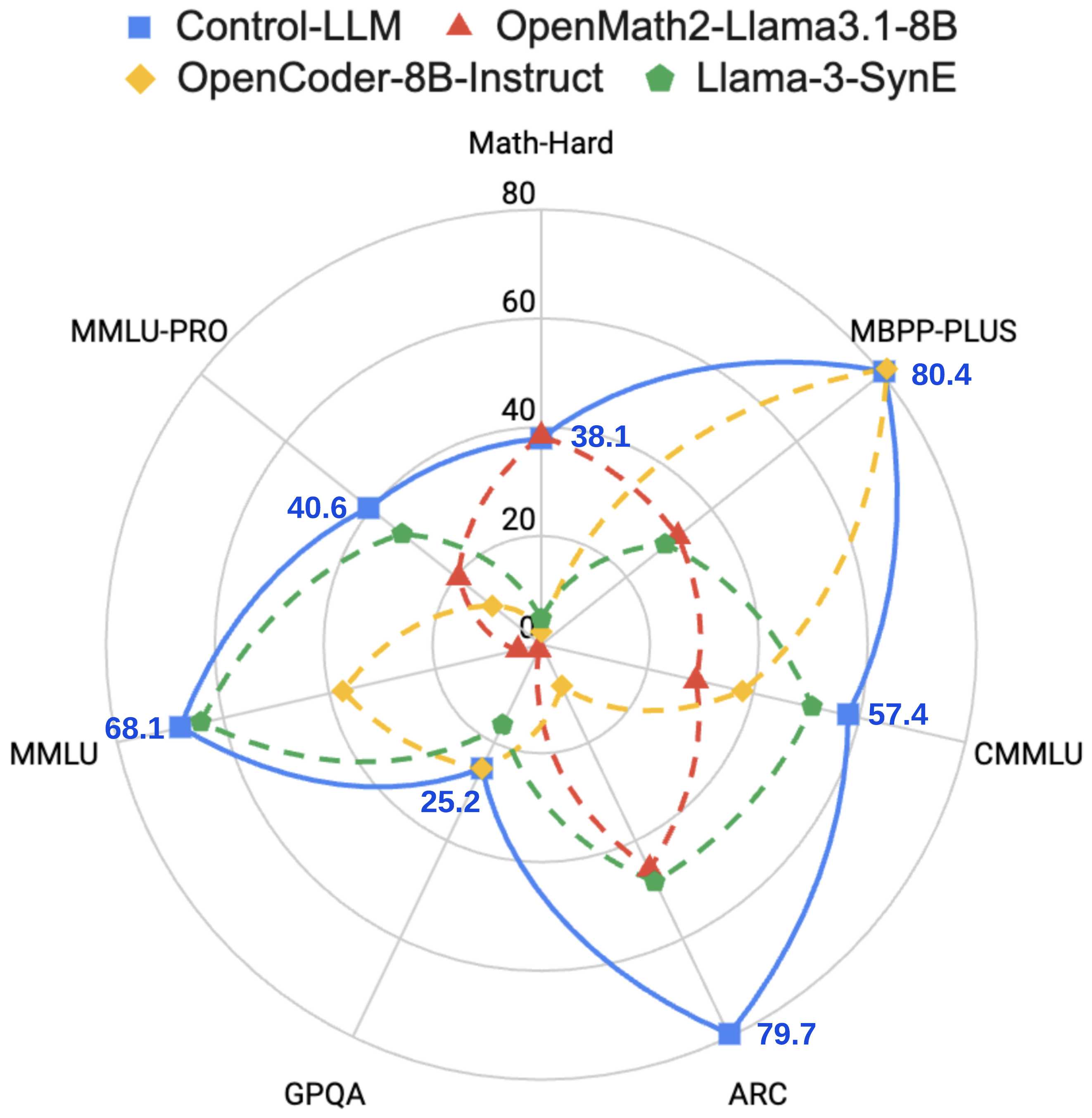}
  \caption{Comparison: Ours vs SOTA Llama-tuned models.}
  \label{fig:control_llm_sota_comparison}
\end{figure}
\vspace{-5pt}


\begin{figure*}[htb]
  \centering
  \begin{subfigure}[t]{0.49\textwidth}
    \centering
    \includegraphics[width=\textwidth]{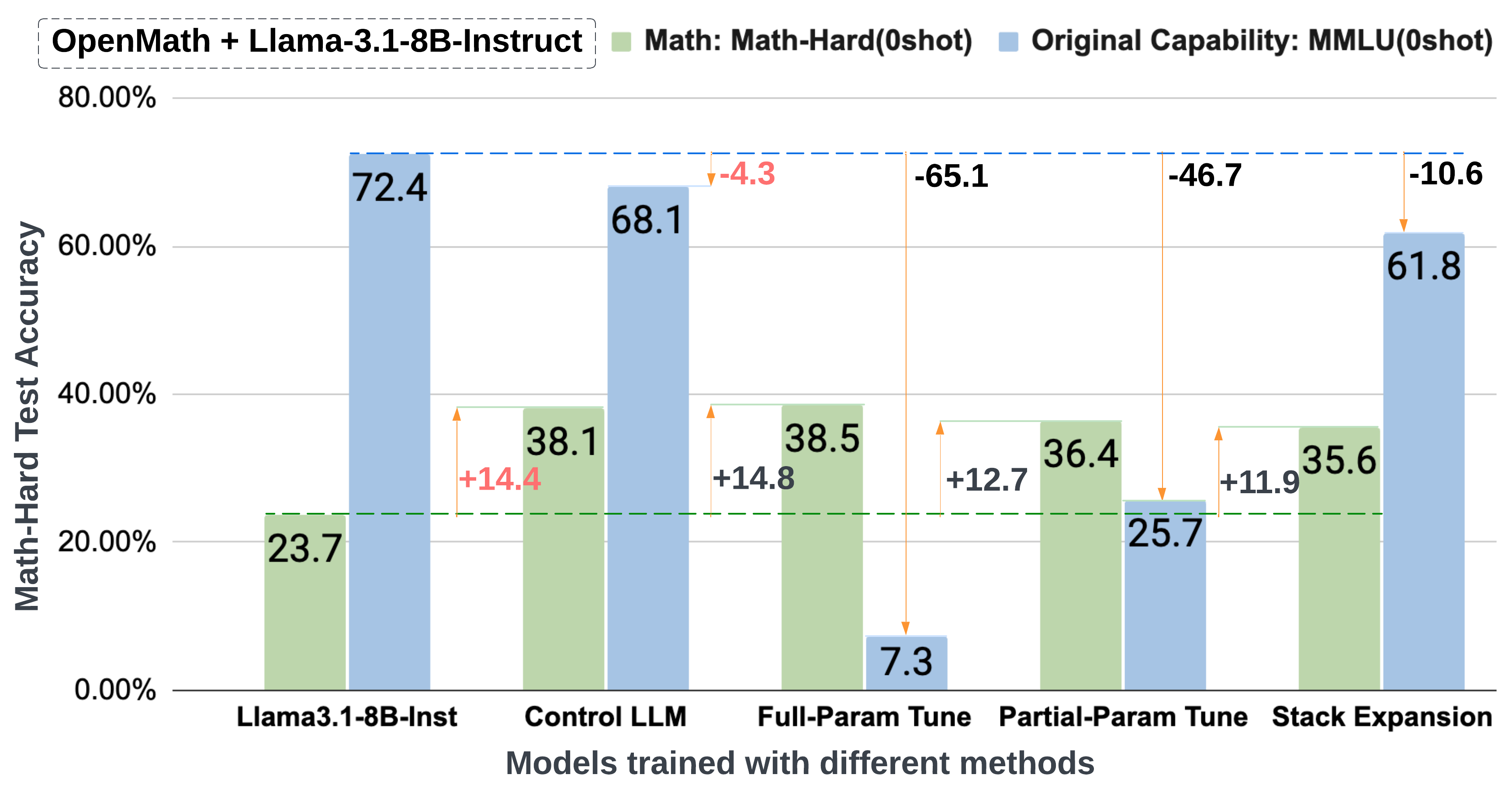}
    \label{fig:catastrophic_forgetting_openmath}
  \end{subfigure}
  \hfill
  \begin{subfigure}[t]{0.49\textwidth}
    \centering
    \includegraphics[width=\textwidth]{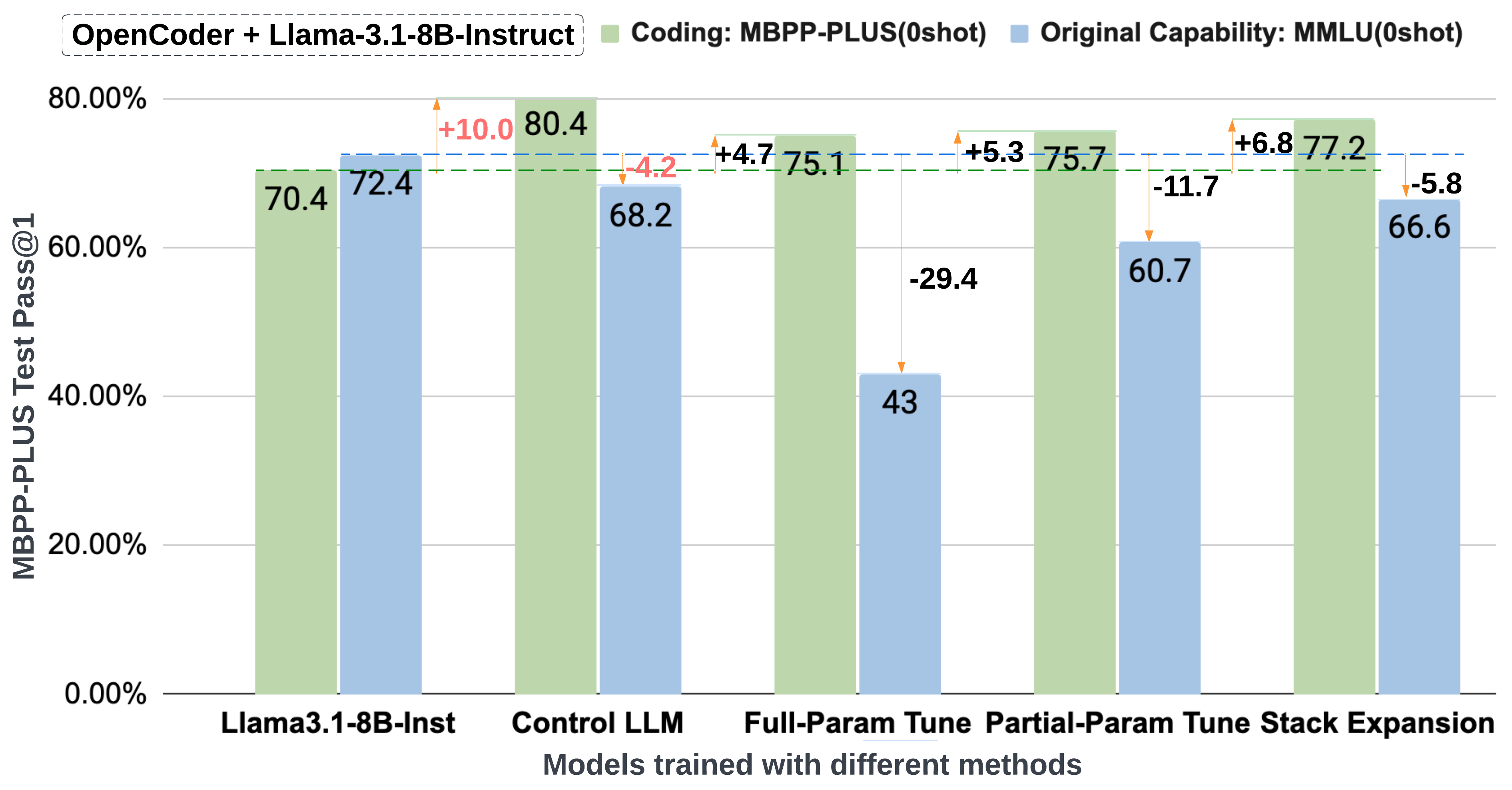}
    \label{fig:catastrophic_forgetting_opencoder}
  \end{subfigure}
  \vspace{-5pt}
  \caption{[\textbf{Result}] Comparison of CF - our method vs others on open-source datasets: \textbf{(left)} OpenMath, \textbf{(right)} OpenCoder.}
  \label{fig:catastrophic_forgetting_combined}
\end{figure*}

\begin{table*}[!htbp]
\centering
\caption{[\textbf{Result}] CSFT on mathematical(0-shot): Open-source models, various CPT approaches, and ControlLLM(Concat-Lerp-MSE).}
\label{tab:csft_results_math}
\begin{tabular}{
  l
  c
  c
  c
  c
  c
  c
  c
  c
  c
  c}
\toprule
\textbf{Model} & \multicolumn{4}{c}{\textbf{Math}} & \multicolumn{5}{c}{\textbf{Original Capabilities}} & \textbf{Avg.}\\
\cmidrule(lr){2-5} \cmidrule(lr){6-10} \cmidrule(lr){11-11}
 & MathHard & Math & GSM8K & Avg. & ARC & GPQA & MMLU & MMLUP & Avg. & \\
\midrule
Llama3.1-8B-Instruct & 23.7 & 50.9 & 85.6 & 52.1 & 83.4 & 29.9 & 72.4 & 46.7 & 60.5 & 56.3 \\
OpenMath2-Llama3.1 & 38.4 & 64.1 & 90.3 & 64.3 & 45.8 & 1.3 & 4.5 & 19.5 & 12.9 & 38.6 \\
\midrule
Full Param Tune & \textbf{38.5} & \textbf{63.7} & 90.2 & \textbf{63.9} & 58.2 & 1.1 & 7.3 & 23.5 & 16.5 & 40.1 \\
Partial Param Tune & 36.4 & 61.4 & 89.0 & 61.8 & 66.2 & 6.0 & 25.7 & 30.9 & 29.3 & 45.6 \\
Stack Expansion & 35.6 & 61.0 & 90.8 & 61.8 & 69.3 & 18.8 & 61.8 & 43.1 & 53.3 & 57.6 \\
Hybrid Expansion & 34.4 & 61.1 & 90.1 & 61.5 & \textbf{81.8} & \textbf{25.9} & 67.2 & \textbf{43.9} & 57.1 & 59.3 \\
\midrule
\textbf{Control LLM*} & 38.1 & 62.7 & \textbf{90.4} & 63.2 & 79.7 & 25.2 & \textbf{68.1} & 43.6 & \textbf{57.2} & \textbf{60.2} \\
\bottomrule
\end{tabular}
\end{table*}

\begin{table*}[!htbp]
\centering
\caption{[\textbf{Result}] CSFT on coding(0-shot). ControlLLM(Concat-Lerp-MSE).
Abbr.(e.g. MBPP+, HE+, MMLUP) in Section~\ref{sec:evaluation}.}
\label{tab:csft_results_coding}
\begin{tabular}{
  l  
  c  
  c  
  c  
  c  
  c  
  c  
  c  
  c  
  c  
  c  
  c} 
\toprule
\textbf{Model} & \multicolumn{5}{c}{\textbf{Coding}} & \multicolumn{5}{c}{\textbf{Original Capabilities}} & \textbf{Avg.}\\
\cmidrule(lr){2-6} \cmidrule(lr){7-11} \cmidrule(lr){12-12}
 & MBPP+ & MBPPS & HE+ & HE & Avg. & ARC & GPQA & MMLU & MMLUP & Avg. & \\
\midrule
Llama3.1-8B-Ins & 70.4 & 67.7 & 66.5 & 70.7 & 69.1 & 83.4 & 29.9 & 72.4 & 46.7 & 60.5 & 64.8 \\
OpenCoder-8B-Ins & 81.2 & 76.3 & 78.0 & 82.3 & 79.5 & 8.2 & 25.4 & 37.4 & 11.3 & 24.6 & 52.1 \\
\midrule
Full Param Tune & 75.1 & 69.6 & 71.3 & 76.8 & 73.3 & 24.4 & 21.9 & 43.0 & 19.2 & 31.5 & 52.4 \\
Partial Param Tune & 75.7 & 71.6 & 74.4 & 79.3 & 75.0 & 70.2 & 28.1 & 60.7 & 32.4 & 48.3 & 61.7 \\
Stack Expansion & 77.2 & 72.8 & 73.2 & 78.7 & 75.6 & 80.0 & 26.3 & 66.6 & 38.2 & 54.2 & 64.9 \\
Hybrid Expansion* & 77.5 & 73.5 & \textbf{76.2} & \textbf{82.3} & 77.1 & 80.9 & \textbf{32.6} & 68.1 & 40.3 & 56.0 & 66.6 \\
\midrule
\textbf{Control LLM*} & \textbf{80.4} & \textbf{75.9} & 74.4 & 81.1 & \textbf{78.3} & \textbf{82.5} & 29.7 & \textbf{68.2} & \textbf{40.9} & \textbf{56.3} & \textbf{67.3} \\
\bottomrule
\end{tabular}
\end{table*}

Large language models (LLMs) achieve emergent capabilities by training on trillions of tokens, requiring \(10^{23}\)--\(10^{25}\) FLOPs \citep{brown2020language, dubey2024llama}. These computational demands make full retraining impractical for many. While pre-trained LLMs provide broad utility, they often lack domain-specific skills. Improving these models via continuous pre-training (CPT) offers a practical alternative to full retraining. Post-training methods—such as supervised fine-tuning (SFT), reinforcement learning from human feedback (RLHF), and model distillation—are further hindered by the limited availability of datasets and methodologies, highlighting the importance of continuous supervised fine-tuning (CSFT).


A primary challenge is \textit{catastrophic forgetting} (CF), where newly acquired knowledge overwrites previously learned information \citep{mccloskey1989catastrophic}. In LLMs, CF is heightened by both large model size and the diverse tasks they must handle.


Existing CF countermeasures include data mixing \citep{sun2019lamol, parmar2024reuse}, regularization (EWC \citep{kirkpatrick2017overcoming}, LwF \citep{li2017learning}), and Parameter-Efficient Fine-Tuning (PEFT) methods (e.g., LoRA \citep{hu2021lora}), yet each has drawbacks. Data mixing is difficult when datasets are undisclosed, regularization can be computationally expensive at large scales, and PEFT may restrict learning capacity \citep{biderman2024lora}.

We propose \textbf{Control LLM}, a method to mitigate CF and integrate new tasks effectively. It employs parallel pre-trained and expanded transformer blocks \citep{zhang2023adding}, aligned via interpolation to balance knowledge retention and new-skill acquisition. Control LLM achieves a \textbf{learn more, forget less} outcome, validated on mathematics (OpenMath2~\citep{toshniwal2024openmathinstruct}), coding (OpenCoder~\citep{opencoder2024}), and multilingual tasks (Llama3-SynE~\citep{survivi2024llama3syne}). Remarkably, it outperforms full-parameter tuning in both learning and CF mitigation for coding and multilingual tasks, while reducing data and compute requirements.

\begin{table*}[!htbp]
\centering
\caption{[\textbf{Result}] CPT on Chinese. Control LLM(Concat-Dlerp). Abbr.: CEvalC(C-Eval-0shot-CoT), CMMLUC(CMMLU-0shot-CoT).}
\label{tab:cpt_results_multilingual}
\begin{tabular}{
  l  
  c  
  c  
  c  
  c  
  c  
  c  
  c  
  c  
  c  
  c} 
\toprule
\textbf{Model} & \multicolumn{5}{c}{\textbf{Chinese}} & \multicolumn{4}{c}{\textbf{Original Capabilities}} & \textbf{Avg.} \\
\cmidrule(lr){2-6} \cmidrule(lr){7-10} \cmidrule(lr){11-11}
 & CEval & CEvalC & CMMLU & CMMLUC & Avg. & BBH & MMLU & MMLUP & Avg. & \\
\midrule
Llama3.1-8B & 48.3 & 12.8 & 51.1 & 14.1 & 13.9 & 65.2 & 65.4 & 35.5 & 45.9 & 29.9 \\
Llama-3-SynE & 57.7 & 22.3 & 57.1 & 22.8 & 22.8 & 61.9 & 64.0 & 32.6 & 42.9 & 32.9 \\
\midrule
Full Param Tune* & 59.0 & 40.2 & \textbf{60.2} & 44.3 & 43.8 & 64.8 & 64.9 & 35.0 & 45.4 & 44.6 \\
Stack Expansion & 56.0 & 32.7 & 55.2 & 33.4 & 33.3 & 62.3 & 65.6 & 35.3 & 44.8 & 39.1 \\
Concat-Lerp* & 57.1 & 34.8 & 57.0 & 37.4 & 37.1 & 64.4 & 64.6 & 35.8 & 45.9 & 41.5 \\
Hybrid Expansion* & \textbf{58.9} & 44.7 & 57.9 & 44.3 & 44.4 & 65.1 & \textbf{65.7} & 36.9 & 46.8 & 45.6 \\
\midrule
\textbf{Control LLM*} & 57.0 & \textbf{44.7} & 56.0 & \textbf{44.9} & \textbf{44.8} & \textbf{68.2} & 65.6 & \textbf{37.9} & \textbf{48.5} & \textbf{46.7} \\
\bottomrule
\end{tabular}
\end{table*}


This approach meets two key production needs: 1) preserving broad capabilities for real-world applications, and 2) enabling high-QPS, low-latency scenarios without large test-time scaling \citep{snell2408scaling}.

\textbf{Contributions} of this paper:
\begin{itemize}[leftmargin=*]
    \item Propose \textbf{Control LLM}, an architecture that mitigates CF by integrating trainable blocks with frozen pre-trained transformer blocks, applicable to both CPT and CSFT.
    \item Highlight the role of hidden-state alignment in retaining prior knowledge while learning new tasks and propose a method to achieve it.
    \item Demonstrate effectiveness across multilingual, mathematical, and coding tasks, achieving gains without degrading existing performance.
\end{itemize}


The remainder of the paper is organized as follows: Section~\ref{related_work} surveys related work, Section~\ref{methodology} details Control LLM, Section~\ref{experiments} outlines experimental settings and results, Section~\ref{discussion_future_work} discusses implications, and concludes the paper with future directions.

\section{Related Work}
\label{related_work}


\textit{Catastrophic forgetting} (CF) describes how new knowledge can overwrite old knowledge in neural networks \citep{mccloskey1989catastrophic}. In LLMs, CF is especially problematic during CPT and supervised fine-tuning, where newly introduced data risks erasing existing capabilities \citep{luo2023empirical}.

\textbf{Regularization-Based Methods}:
Elastic Weight Consolidation (EWC) \citep{kirkpatrick2017overcoming} and its variants (Synaptic Intelligence \citep{zenke2017continual}, Memory Aware Synapses \citep{aljundi2018memory}) constrain updates for parameters deemed critical to past tasks.

\textbf{Replay Methods}:
Experience Replay \citep{rolnick2019experience} and Generative Replay \citep{shin2017continual, sun2019lamol, huang2024mitigating} reintroduce earlier tasks via real or synthetic data. Curated mixtures further reduce distribution drift \citep{parmar2024reuse, xi2024practice, wu2024llama}.

\textbf{Parameter Isolation Methods}:
Progressive Neural Networks \citep{rusu2016progressive} freeze existing modules and append new ones. PackNet \citep{mallya2018packnet} prunes parameters for reuse. PEFT (e.g., LoRA \citep{hu2021lora}, prefix-tuning \citep{li2021prefix}, LLaMA-Adapter \citep{zhang2023llama}) fine-tunes a small subset of parameters, reducing computation but limiting learning capacity.

\textbf{Knowledge Distillation}:
Learning without Forgetting (LwF) \citep{li2017learning} and Selective Distillation \citep{yu2025select} transfer knowledge from older or parallel teachers to preserve prior tasks.

\textbf{Architecture-Based Methods}:
Adapter modules \citep{houlsby2019parameter} insert task-specific layers, leaving the original model intact. LLaMA Pro \citep{wu2024llama} and SOLAR \citep{kim2023solar} expand model depth through additional transformer blocks or Depth Up-Scaling (DUS). Llama-MoE v2 \citep{qu2024llama} does the model upcycling by partitioning FFN or Attention layers into sparse experts and trains top-k gate.


\begin{figure*}[htb]
  \centering
  \includegraphics[width=1.0\textwidth]{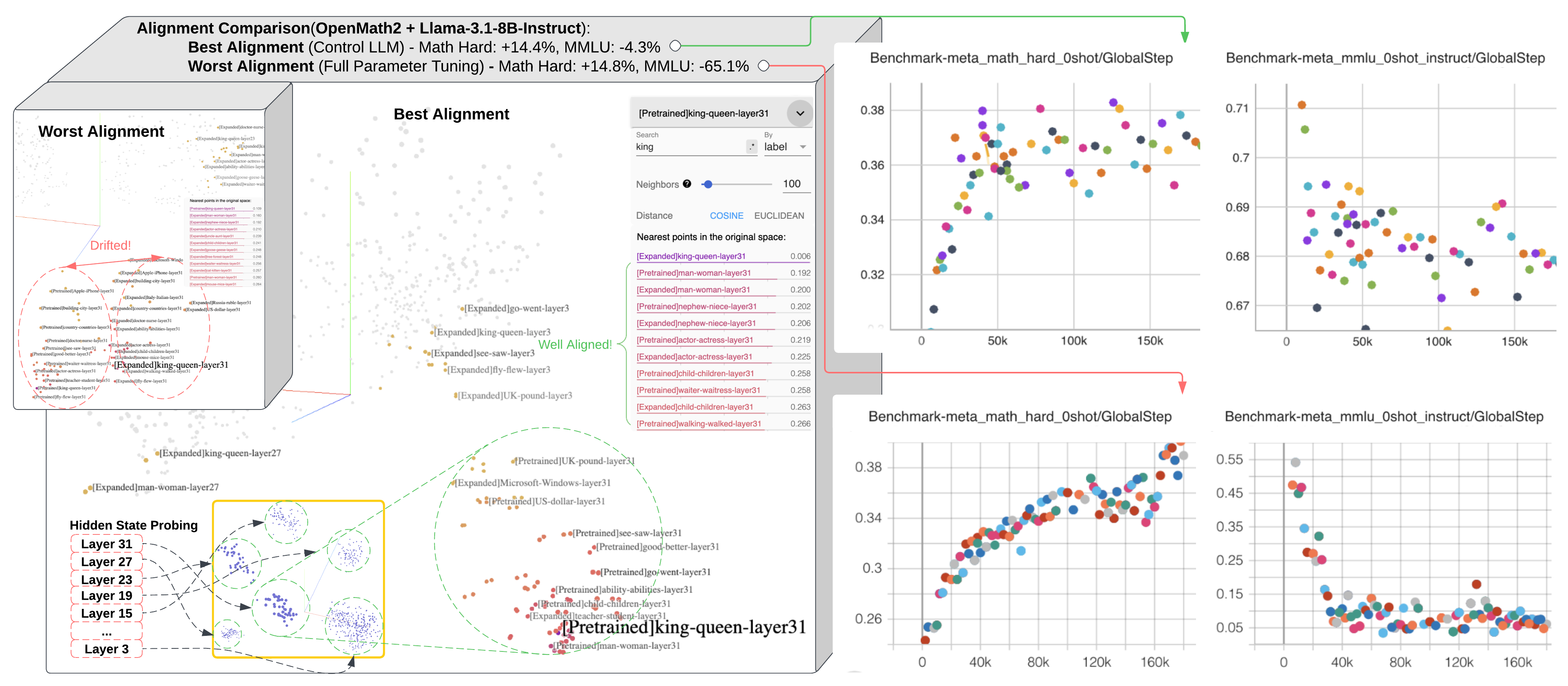}
  \vspace{-5pt}
  \caption{[\textbf{Why}] Hidden State Alignment Comparison: \textbf{Best Alignment(Control LLM)} vs \textbf{Worst Alignment(Full-Parameter Tuning)}.}

  \label{fig:alignment_comparison}
\end{figure*}


\textbf{Control LLM} combines an architecture-based model upcycling approach with parameter isolation. In addition, it aligns hidden states of parallel pre-trained and expanded blocks via interpolation and divergence loss, to the best of our knowledge, no prior work has explored. Unlike many PEFT methods, it does not curtail learning capacity or demand original training data. As a result, it retains previously learned skills while accommodating new tasks.

\section{Methodology}
\label{methodology}


This section introduces the Control LLM approach. We first motivate the need for hidden-state alignment (Section~\ref{sec:alignment}), then present our architecture (Section~\ref{sec:control_arch}) and training procedure (Section~\ref{sec:training}).


\subsection{Hidden-State Alignment in Transformer Layers}
\label{sec:alignment}



Transformer models \citep{vaswani2017attention} process tokens through layers, producing hidden-states that encode learned representations. Preserving alignment in these hidden-states is critical to mitigating \textit{catastrophic forgetting} (CF), where new data overwrites prior knowledge \citep{ramasesh2020anatomy}.

Figure~\ref{fig:alignment_comparison} (more in Appendix~\ref{sec:probing_result}) demonstrates the impact of hidden-state alignment during Continuous Pre-training (CPT) and Continuous Supervised Fine-Tuning (CSFT). Probing with sentences like “king is to queen” (Appendix~\ref{sec:probing_data}) reveals that well-aligned hidden states between [Expanded] (tuned) and [Pretrained] models preserve semantic relationships, improving downstream performance. In math tasks (OpenMath2~\citep{nvidia2024openmathinstruct2}), model tuned with alignment improves MathHard accuracy by +14.4\% and limited MMLU degradation to -4.3\%. Conversely, misalignment caused severe performance drops (e.g., MMLU fell from 72.4\% to 7.3\%). Benchmark results show that alignment sustains MMLU performance over 150K training steps, whereas misaligned tuning leads to degradation within 40K steps. Similar benefits were observed in fine tuning coding (OpenCoder~\citep{opencoder2024}) and multilingual (Llama3-SynE~\citep{survivi2024llama3syne}) tasks.


Key observations include:
\vspace{-5pt}
\begin{enumerate}[leftmargin=*]
    \item \textbf{Semantic Stability}: Hidden states of analogous sentences (e.g., king:queen, man:woman) exhibit strong cosine and Euclidean similarity.
    \item \textbf{Layer Clustering}: Analogous sentences show high similarity within layers and distinct clustering across layers.
    \item \textbf{CF Mitigation}: Aligned [Expanded] and [Pretrained] states reduce destructive interference and CF.
\end{enumerate}


These results demonstrate that coherent transformations across layers enable models to refine rather than overwrite prior knowledge—a critical requirement for large, deep LLMs. The following sections explore how Control LLM exploits these insights to mitigate CF effectively.

\begin{figure*}[ht]
    \centering
    \includegraphics[width=6.51in]{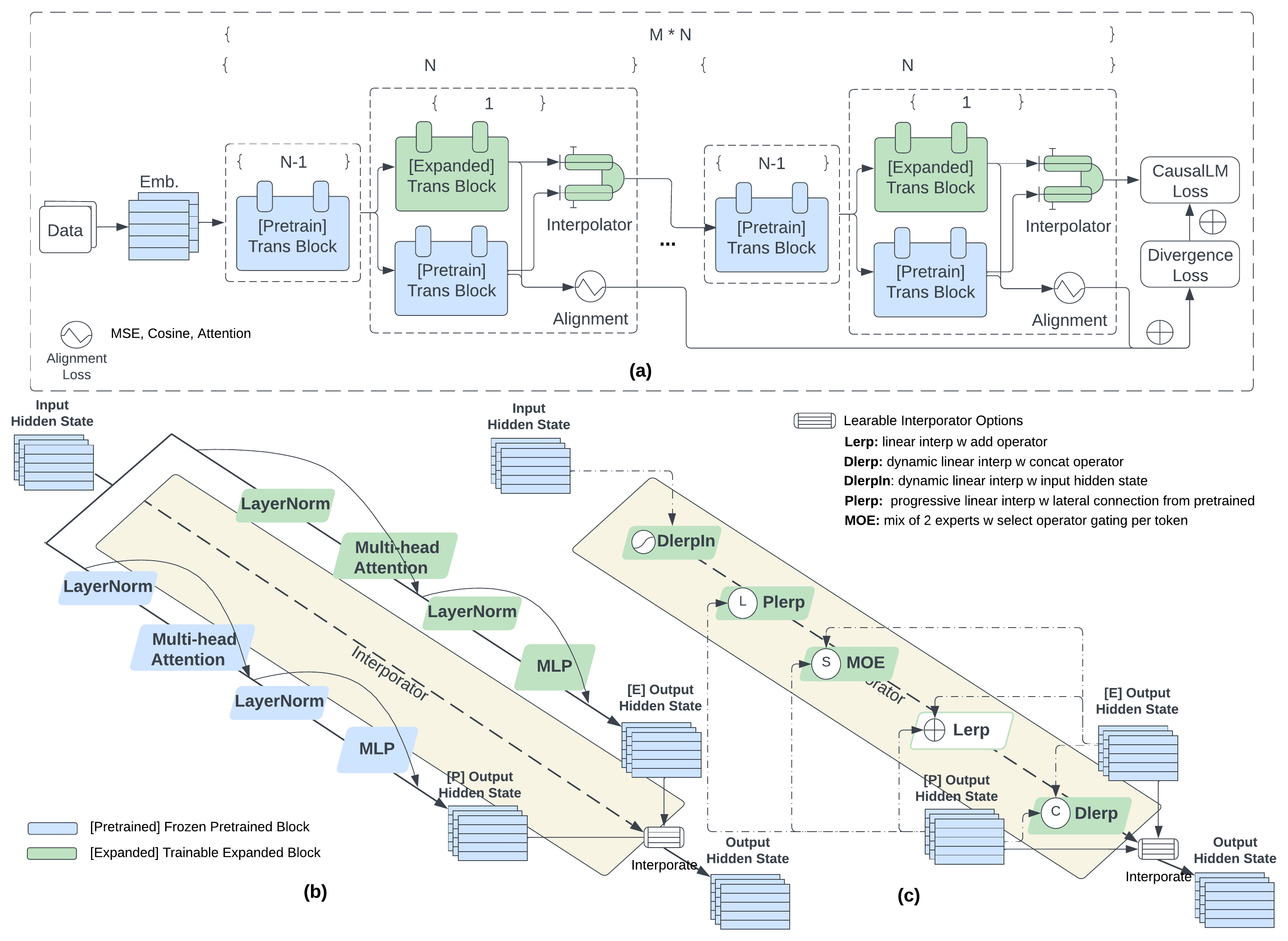} 
    \vspace{-5pt}
    \caption{[\textbf{How}] Control LLM Architecture. \textbf{(a)} Expanded blocks added every \(N-1\) layers connect to frozen blocks via interpolators. \textbf{(b)} Interpolators align hidden-states to produce final representations. \textbf{(c)} Different interpolation strategies are explored.}
    \label{fig:control_llm_architecture}
\end{figure*}

\subsection{Control LLM Architecture}
\label{sec:control_arch}


As illustrated in Figure~\ref{fig:control_llm_architecture}, Control LLM augments every \((N-1)\)th transformer layer (with \(N\) as a hyperparameter) by creating two branches:


\begin{enumerate}[leftmargin=*]
    \item \textbf{Pre-trained Transformer Block}: The original, frozen block that retains established capabilities.
    \item \textbf{Expanded Transformer Block}: A trainable copy that acquires new knowledge or skills from additional data.
\end{enumerate}


Interpolation with an alignment mechanism fuses these branches’ outputs, integrating new representations with existing knowledge, with \textbf{no initial training performance decrease} entailed by changing a trained network’s structure.

\subsection{Alignment Mechanisms and Interpolation Strategies}
\label{sec:interp_strategies}


We fuse \(\{h_{\text{pre-trained}}, h_{\text{expanded}}\}\) through various interpolation methods, allowing to apply a layer wise divergence loss to maintain consistent alignment:


\paragraph{Linear Interpolation (Lerp).}


A scalar \(\alpha \in [0,1]\) mixes the hidden-states:
\[
h_{\text{combined}} = (1-\alpha) h_{\text{pre-trained}} + \alpha\,h_{\text{expanded}}.
\]

\paragraph{Dynamic Linear Interpolation (Dlerp).}



Here, \(\alpha\) depends on concatenated outputs from both branches:
\[
\alpha(x) = \sigma\bigl(W[x_{\text{pre-trained}}, x_{\text{expanded}}] + b\bigr),
\]
allowing \(\alpha\) to vary \emph{per token}. Dlerp can address CF without explicit divergence loss.

\paragraph{DlerpIn.}



Instead of conditioning \(\alpha\) on layer outputs, DlerpIn uses input hidden-states:
\[
\alpha(h_{\text{input}}) = \sigma\bigl(W_{\text{in}}\,h_{\text{input}} + b_{\text{in}}\bigr).
\]
This leverages contextual clues before transformation.

\paragraph{Mixture of Experts (MoE) Gating.}

Inspired by Mixture of Experts, a gating network \(G\) predicts a probability distribution over pre-trained (\(h_{\text{pre-trained}}\)) and expanded (\(h_{\text{expanded}}\)) representations using input hidden-states \(h_{\text{input}}\), where \([\alpha_{\text{pre-trained}}, \alpha_{\text{expanded}}] = \text{softmax}(W_g h_{\text{input}} + b_g)\) and \(b_g\) is a bias vector. The top expert per token is selected via \(\text{argmax}\):
\[
h_{\text{combined}} = 
\begin{cases}
h_{\text{pre-trained}}, & \text{if pre-trained is chosen},\\
h_{\text{expanded}}, & \text{if expanded is chosen.}
\end{cases}
\]
Unlike continuous blending (e.g., DlerpIn), this "hard" interpolation enforces a discrete choice, which can amplify catastrophic forgetting (CF).

\vspace{-5pt} 
\paragraph{Progressive Linear Interpolation (Plerp).}




A lateral layer (near-identity) transforms \( h_{\text{pre-trained}} \) prior to blending: 
\( h_{\text{lateral}} = W_{\text{lateral}}\,h_{\text{pre-trained}} \) and 
\( h_{\text{combined}} = (1-\alpha)\,h_{\text{lateral}} + \alpha\,h_{\text{expanded}} \).
Gradual updates let the expanded layer integrate new features while preserving original representations.

\subsubsection{Divergence Loss for Alignment.}

The divergence loss penalizes drift between \(\{h_{\text{pre-trained}}, h_{\text{expanded}}\}\), averaged over layers:
\[
\mathcal{L}_{\text{divergence}} = \frac{1}{L} \sum_{l=1}^{L} \mathbb{E}_{i,j}\bigl[\alpha^{(l)}(i,j)\,\Delta(h_{\text{pre-trained}}^{(l,i,j)}, h_{\text{expanded}}^{(l,i,j)})\bigr],
\]
where \(L\) is the number of layers expanded, \(i\) indexes over the batch, \(j\) over sequence positions, and \(l\) over layers. \(\Delta\) can be cosine, MSE, or attention divergence. Scaling by \(\alpha^{(l)}\) enables selective alignment for each layer. Empirically, MSE and cosine work well by constraining both direction and magnitude (details in Section~\ref{ablation_study}).

\subsubsection{Model Expansion Strategies}
\label{structure_analysis}





\begin{figure}
    \centering
    \includegraphics[width=3.2in]{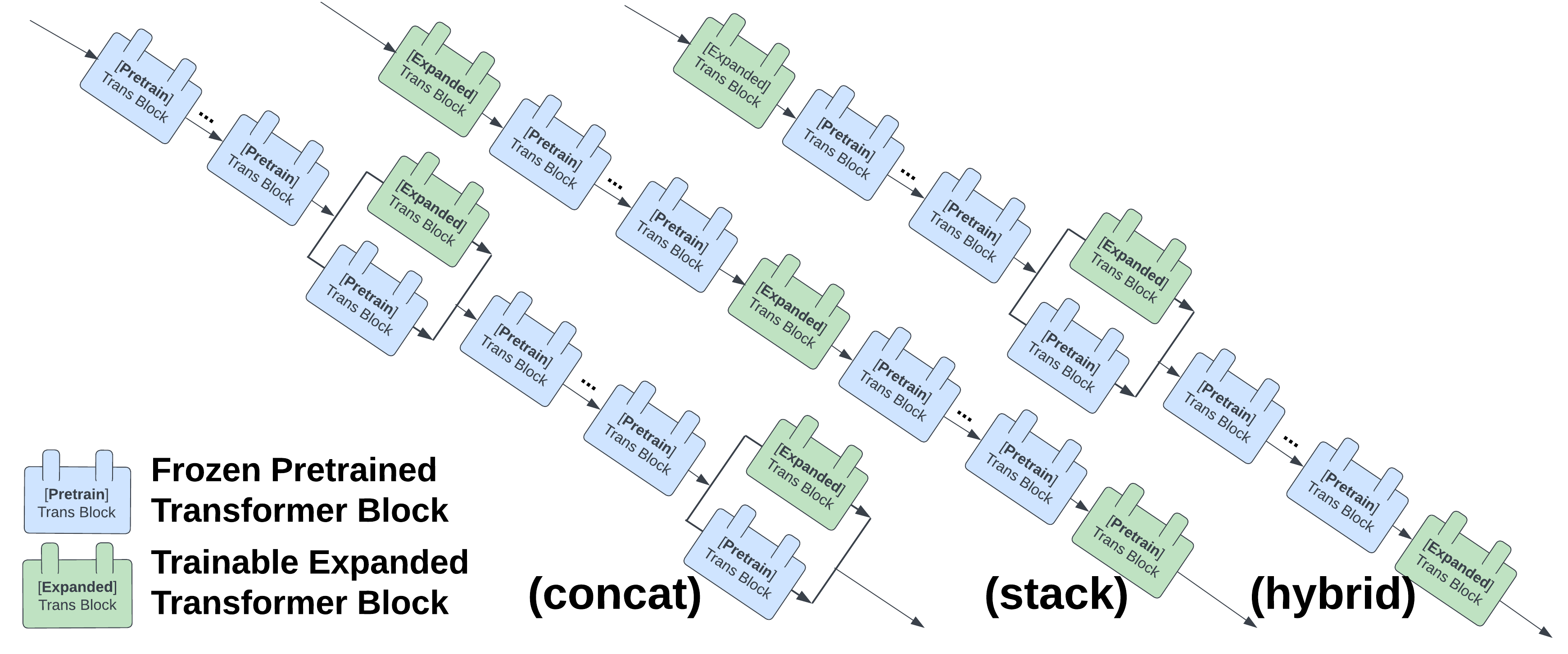}
    \vspace{-5pt}
    \caption{[\textbf{How}] Structure analysis: \textbf{(concat)} the default dual structure. \textbf{(stack)} stack the expanded block following LLaMA Pro. \textbf{(hybrid)} hybrid structure of concat and stack.}
    \label{fig:control_llm_structure_analysis}
    \vspace{-6pt}
\end{figure}

Figure~\ref{fig:control_llm_structure_analysis} illustrates three expansion strategies for each \((N-1)\)th layer:
\begin{itemize}[leftmargin=*]
    \item \textbf{Concat}: Adds a parallel “side-car” for dual-branch blending, initialized as a copy of pretrained transformer block.
    \item \textbf{Stack}: Stacks copied new layers (adapted from \citet{wu2024llama}) above the original, initialized by zeroing out specific projections(o\_proj, down\_proj) to approximate an identity mapping.
    \item \textbf{Hybrid}: Alternates between stack and concat.
\end{itemize}




Ablation studies show that \textbf{concat} balances learning and retention best (Table~\ref{tab:ablation_study}); hence we adopt it as our default.

\subsection{Training Procedure}
\label{sec:training}

\begin{enumerate}[leftmargin=*]
    \item \textbf{Initialization}: Duplicate pre-trained blocks to form expanded blocks, preserving initial alignment.
    \item \textbf{Data Preparation}: Use task-specific datasets. CPT employs sequence packing (8K context), while CSFT uses up to 132K context, supervising response tokens.
    \item \textbf{Alignment Training}:
    \[
    \mathcal{L} = \mathcal{L}_{\text{task}} + \lambda\,\mathcal{L}_{\text{divergence}},
    \]
    where \(\mathcal{L}_{\text{task}}\) denotes cross-entropy, \(\mathcal{L}_{\text{divergence}}\) enforces hidden-state alignment, and \(\lambda\) is a hyperparameter.
    \item \textbf{Optimization}: Train only expanded layers and interpolator. Use AdamW \citep{loshchilov2017decoupled} with a weight decay set to 10\% of the learning rate, a cosine scheduler (1000 warm-up, peak \(5\times10^{-5}\), min \(1\times10^{-5}\)). For CPT, batch size is 8*64 globally; for CSFT, 32*8 for math, 12*8 for coding.

    \item \textbf{Model Selection}: Train for 2--6 epochs, reserving 20K validation samples. Pick best-performing checkpoint on new tasks to ensure fairness across methods with varying convergence speeds. For coding tasks, pick the best from phase 1 to train phase 2.

\end{enumerate}


\section{Experiments}
\label{experiments}
We evaluate Control LLM under both Continuous Pre-training (CPT) and Continuous Supervised Fine-Tuning (CSFT), focusing on four research questions:


\begin{itemize}[leftmargin=*]
    \item \textbf{RQ1}: Can Control LLM effectively learn new tasks while mitigating CF in both CPT and CSFT?
    \item \textbf{RQ2}: Does Control LLM generalize to a broad range of tasks (multilingual, mathematical reasoning, coding, and out-of-domain)?
    \item \textbf{RQ3}: How does it perform relative to baseline methods and open-source state-of-the-art (SOTA) models from the same base?
    \item \textbf{RQ4}: Which Control LLM configurations yield the best performance and why?
\end{itemize}


Benchmarks in Tables~\ref{tab:csft_results_math}, \ref{tab:csft_results_coding}, \ref{tab:cpt_results_multilingual}, and \ref{tab:ablation_study} address these questions across mathematical, coding, multilingual, and general capabilities.

\subsection{Experimental Setup}
\label{sec:exp_setup}

\begin{table}[ht]
\centering
\begin{tabular}{@{\hskip 5pt}l@{\hskip 5pt}l@{\hskip 5pt}c@{\hskip 5pt}c@{\hskip 5pt}}
\toprule
\textbf{Dataset}           & \textbf{Task}      & \textbf{Samples} & \textbf{Tokens} \\
\midrule
Llama3-SynE (-EN)          & Multilingual      & 47M              & 35.8B         \\
OpenMath2         & Mathematics       & 13.27M           & 5.1B          \\
OpenCoder-SFT-Phase1       & Coding            & 4.02M            & 2.4B          \\
OpenCoder-SFT-Phase2       & Coding            & 0.43M            & 245.4M        \\
\bottomrule
\end{tabular}
\caption{[\textbf{What}] Summary of datasets used in experiments, categorized by task type, with sample and token counts.}
\label{tab:datasets}
\end{table}

\subsubsection{Datasets}
\label{datasets}

Table~\ref{tab:datasets} summarizes the open-source datasets used to validate Control LLM.


\textbf{CPT (Llama3-SynE \citep{survivi2024llama3syne}):} 
Combines English/Chinese corpora to enhance multilingual proficiency. To demonstrate CF mitigation without data replay, we omit English subsets (\texttt{book\_en}, \texttt{encyclopedia\_en}, \texttt{qa\_forum\_en}), retaining only Chinese, code, and math data (\texttt{book\_cn}, \texttt{encyclopedia\_cn}, \texttt{qa\_forum\_cn}, \texttt{web\_cn}, \texttt{code\_en}, \texttt{math\_en}, \texttt{synthesis\_en}).


\textbf{CSFT:} 
\begin{itemize}[leftmargin=*]
    \item \textbf{OpenMath2} \citep{nvidia2024openmathinstruct2}: 14M math samples, filtered sequences longer than 1024 Llama tokens to 13.27M. 
    \item \textbf{OpenCoder} \citep{opencoder2024}: We use SFT-Phase1 and SFT-Phase2 data exclusively for CSFT reducing training data from 2.5T to 2.6G tokens.
\end{itemize}

\subsubsection{Baselines}



We compare Control LLM to:
\begin{itemize}[leftmargin=*]
    \item \textbf{Full Parameter Tuning}: All parameters are trainable.
    \item \textbf{Partial Parameter Tuning}: Freezes all but every \((N-1)\)th transformer layer.
    \item \textbf{Stack Expansion}: Following LLaMA Pro \citep{wu2024llama}, new layers are stacked while original layers remain frozen (\textbf{stack} strategy).
\end{itemize}


We also compare to open-source models trained on similar datasets and same base model:
\begin{itemize}[leftmargin=*]
    \item \textbf{Llama-3-SynE} \citep{wu2024llama}
    \item \textbf{OpenMath2-Llama3.1-8B} \citep{nvidia2024openmath}
    \item \textbf{OpenCoder-8B-Instruct} \citep{infly2024opencoder}
\end{itemize}

\subsubsection{Evaluation}
\label{sec:evaluation}


We evaluate our models on a set of general benchmarks by implementing a customized version of lm-eval-harvness\citep{eleutherai2024evaluation}. Abbreviations (e.g., MBPP+, HE+) appear in Tables~\ref{tab:csft_results_math}--\ref{tab:cpt_results_multilingual} for brevity.

\begin{itemize}[leftmargin=*]
    \item \textbf{Math}: Exact match accuracy of Math-0shot (Math), Math-Hard-0shot (MathH), and GSM8K (G8K).
    \item \textbf{Coding}: Pass@1 of MBPP-Sanitize-0shot (MBPPS), MBPP-Plus-0shot (MBPP+), HumanEval-Greedy (HE), and HumanEval-Plus-Greedy (HE+).
    \item \textbf{Chinese}: Evaluation includes C-Eval-0shot (CEval), C-Eval-0shot-CoT (CEvalC), CMMLU-0shot (CMMLU), and CMMLU-0shot-CoT (CMMLUC). CEvalC and CMMLUC specifically assess chain-of-thought instruction following in Chinese (details in Appendix~\ref{sec:eval_details}). Note: CEval uses the validation split, as test split ground truth is unavailable.
    \item \textbf{Original Capabilities}: 
        \begin{itemize}
            \item \textbf{CSFT}: ARC\_Challenge-0shot (ARC), GPQA-0shot (GPQA), MMLU-0shot (MMLU), and MMLU\_Pro-5shot (MMLUP).
            \item \textbf{CPT}: BBH-3shot (BBH), MMLU-5shot (MMLU), and MMLU\_Pro-5shot (MMLUP).
        \end{itemize}
\end{itemize}


CSFT benchmarks utilize a maximum model length of 8192, while CPT employs 5192. Results are reported as size-weighted averages within each task group, along with the overall mean of these group averages.




\subsection{Results}
\label{results}


We report Control LLM in two settings—\textbf{Hybrid Expansion} and \textbf{Concat Expansion} (referred to simply as Control LLM)—across math, coding, and multilingual tasks in Tables~\ref{tab:csft_results_math}, \ref{tab:csft_results_coding}, and \ref{tab:cpt_results_multilingual}.

\subsubsection{Continuous Supervised Fine-tuning (CSFT)}
\label{results_csft}

\paragraph{Math (Table~\ref{tab:csft_results_math}).}
OpenMath2-Llama3.1-8B~\citep{nvidia2024openmath} fine-tunes the Llama3.1-8B \emph{base} with 13.27M math samples, showing strong task-specific gains. Control LLM (from Llama3.1-8B-Instruct) matches these gains while preserving Original Capabilities. OpenMath2 employs checkpoint averaging (+2\%), but we report single-checkpoint results for reproducibility. Despite effectively learning math, both OpenMath2 and Full-Parameter Tuning degrade Original Capabilities (e.g., MMLU drops from 72.4\% to $<\!10\%$), whereas Control LLM remains robust even after extended training (Appendix~\ref{sec:eval_details}). Stack Expansion~\citep{wu2024llama} and Partial-Parameter Tuning mitigate catastrophic forgetting, but \textbf{Control LLM-Hybrid} outperforms these approaches. Our final reported results use \textbf{Control LLM-Concat}.

\paragraph{Coding (Table~\ref{tab:csft_results_coding}).}
Control LLM rivals OpenCoder-8B-Instruct~\citep{infly2024opencoder} while using $\sim\!1000\times$ fewer tokens (2.6G vs.\ 2.5T) and no CPT, and it exceeds OpenCoder-8B-Instruct on Original Capabilities. By contrast, OpenMath2-Llama3.1 (CPT+SFT) obtains a +10.3\% coding improvement but reduces Original Capabilities from 60.5\% to 24.6\%. Even extensive hyper-parameter tuning for Full-Parameter Tuning yields only +4.2\% while continually eroding prior knowledge. Stack Expansion and Partial-Parameter Tuning (8 blocks) alleviate forgetting, but \textbf{Control LLM-Hybrid} outperforms both and nearly matches \textbf{Control LLM-Concat}. We open-source both expansions.

\subsubsection{Continuous Pre-Training
(CPT)}
\label{results_cpt}

\paragraph{Multilingual (Table~\ref{tab:cpt_results_multilingual}).} By excluding English data (100B vs.\ 35.8G tokens), Control LLM rivals Llama-3-SynE~\citep{wu2024llama}, which relies on data mixture/replay. Moreover, Control LLM improves CEvalC and CMMLUC (CoT in Chinese) by +31.9\% and +30.8\%, respectively, while preserving Original Capabilities. It thus surpasses replay-based methods (Llama-3-SynE), Full-Parameter Tuning, Stack Expansion~\citep{wu2024llama}, and even the base model. Observations reveal: CPT shows less CF than CSFT, with Hybrid Expansion excelling in deeper architectures. Control LLM not only mitigates forgetting and improves prior competencies for CPT.



\subsubsection{Ablation Studies}
\label{ablation_study}

\begin{table}[ht]
\centering
\caption{[\textbf{Where}] Ablation Study. Abbr. MathH(Math Hard), Cos(Cosine), G8K(GSM8K)}
\label{tab:ablation_study}
\begin{tabular}{@{\hskip 4pt}l@{\hskip 4pt}cc@{\hskip 4pt}cc@{\hskip 4pt}c@{\hskip 4pt}}
\toprule
\textbf{Model} & \multicolumn{2}{c}{\textbf{Math}} & \multicolumn{2}{c}{\textbf{Original}} & \textbf{Avg.} \\
\cmidrule(lr){2-3} \cmidrule(lr){4-5} \cmidrule(lr){6-6}
 & \textbf{MathH} & \textbf{G8K} & \textbf{GPQA} & \textbf{MMLU} & \textbf{} \\
\midrule
Full-Param & \textbf{38.5} & 90.2 & 1.1 & 7.3 & 34.3 \\
\midrule
Lerp8 & 36.1 & 90.2 & 18.5 & 58.8 & 50.9 \\
Lerp8-MSE & 35.9 & 90.1 & 28.4 & \textbf{71.7} & \textbf{56.5} \\
Lerp8-Cos & 34.9 & 89.2 & 29.5 & 71.0 & 56.2 \\
Dlerp8 & \textbf{36.7} & 90.0 & \textbf{28.8} & 69.1 & 56.2 \\
Dlerp8-MSE & 35.6 & 89.5 & 19.9 & 66.1 & 52.8 \\
Dlerp8-Cos & 36.5 & \textbf{90.2} & 26.1 & 69.1 & 55.5 \\
\midrule
DlerpIn8 & 34.1 & 87.6 & \textbf{25.7} & \textbf{68.2} & \textbf{53.9} \\
Plerp8 & \textbf{36.5} & \textbf{90.9} & 9.2 & 41.9 & 44.6 \\
MoE8 & 34.1 & 89.7 & 20.5 & 63.3 & 51.9 \\
\midrule
Lerp16-MSE & 38.1 & 90.4 & \textbf{25.2} & \textbf{68.1} & \textbf{55.5} \\
Dlerp16 & 37.7 & \textbf{91.1} & 22.3 & 64.2 & 53.8 \\
Dlerp32 & \textbf{38.6} & 91.4 & 22.1 & 60.9 & 53.3 \\
\midrule
Lerp8MSE0 & \textbf{35.8} & \textbf{90.6} & 7.4 & 53.4 & 46.8 \\
Lerp8MSE0* & 35.1 & 89.3 & 31.3 & \textbf{72.4} & \textbf{57.0} \\
Lerp8MSE0*M & 33.9 & 89.1 & \textbf{32.4} & 72.2 & 56.9 \\
\midrule
Lerp8MSE\(\alpha\) & 32.6 & 87.3 & 29.8 & 71.8 & 55.4 \\
\bottomrule
\end{tabular}
\end{table}

\begin{itemize}[leftmargin=*]

    \item \textbf{Control LLM Architecture}: 
    Tables~\ref{tab:csft_results_math}--\ref{tab:cpt_results_multilingual} compare stack, partial-parameter tuning, and our expansions. Control LLM-Hybrid and Control LLM-Concat (Section~\ref{structure_analysis}) surpass stack and partial tuning in mitigating CF.

    \item \textbf{Interpolation Strategies}: 
    Table~\ref{tab:ablation_study} assesses Lerp, Dlerp, DlerpIn, Plerp, and MoE on OpenMath2~~\citep{toshniwal2024openmathinstruct}. 
    Lerp/Dlerp balance CF mitigation and new-task performance best. DlerpIn excels at CF mitigation but learns fewer new skills, while Plerp does the inverse. DlerpIn outperforms MoE (“hard” interpolation) by “soft” methods.

    \item \textbf{Divergence Loss}: 
    Removing divergence loss results in more forgetting (though still better than full-parameter tuning and stack expansion), emphasizing its importance. Dlerp remains robust without it, thanks to gradual, soft interpolation (Appendix~\ref{sec:ablation_study}). MSE generally works well, constraining direction and magnitude. Lerp8-Cosine mitigates CF initially but deteriorates with prolonged training.

    \item \textbf{Number of Layers}: 
    Table~\ref{tab:ablation_study} shows that expanding 16 or 32 layers (vs.\ 8) improves performance without severely impacting Original Capabilities. Sixteen layers match full-parameter tuning in MathH; we open-source this configuration. Eight layers remain ideal for production due to the performance–cost trade-off.

    \item \textbf{Interpolation Mechanisms}: 
    Interpolation proves essential; merely aligning hidden-states (e.g., Table~\ref{tab:ablation_study}/Lerp8MSE0: \(\alpha=0\) plus MSE) does not fully prevent CF. However, training with \(\alpha=0\) and MSE, then inferring at \(\alpha=0.5\) (Table~\ref{tab:ablation_study}/Lerp8MSE0*), effectively mitigates CF, confirming the synergy of alignment \emph{and} interpolation. Merging pre-trained and expanded blocks (e.g., Lerp8MSE0*M) degrades performance, likely due to transformer nonlinearity, warranting further research.

    \item \textbf{Fixed vs.\ Learnable \(\alpha\)}: 
    For Lerp, a fixed \(\alpha=0.5\) provides good CF mitigation and faster convergence. Making \(\alpha\) learnable slows convergence (Table~\ref{tab:ablation_study}/Lerp8MSE\(\alpha\)), similar to Dlerp’s bias term. Consequently, all final results freeze \(\alpha\) or relevant bias parameters.
\end{itemize}

\section{Discussion, Future Work, and Conclusion}
\label{discussion_future_work}

Control LLM presents a more effective model upcycling architecture compared to MoE and other baselines. It leverages a continuous and learnable interpolation mechanism, eliminating hard gating and expert silos, enables effective hidden-state alignment and seamless knowledge integration, making it particularly well-suited for dynamic adaptation to evolving data distributions. However, it comes with the trade-off of increasing the number of activated parameters.

Future work includes:
\begin{itemize}[leftmargin=*]
    \item Evolve the architecture to enable effective weight fusion.
    \item Extend to multi-modality and RLHF.
    \item Explore theoretical foundations of hidden-state alignment.
\end{itemize}

In summary, \textbf{Control LLM} offers a novel solution to mitigate \textit{catastrophic forgetting} (CF) in large language models by employing interpolation-based alignment between pre-trained and expanded representations. It achieves a robust balance between retaining prior knowledge and acquiring new skills, consistently outperforming baseline methods with much less data and computation. The approach stabilizes fine-tuning, enabling smooth integration of new knowledge without degrading existing performance, while being robust to hyper-parameter sensitivity and avoiding loss spikes during training (Appendix~\ref{sec:train_details}).





\bibliographystyle{plainnat}

\newpage
\appendix
\onecolumn

\section{The effectiveness of Control LLM in addressing Catastrophic Forgetting}
\label{sec:train_details}

\begin{figure*}[ht]
    \centering
    \includegraphics[width=6.51in]{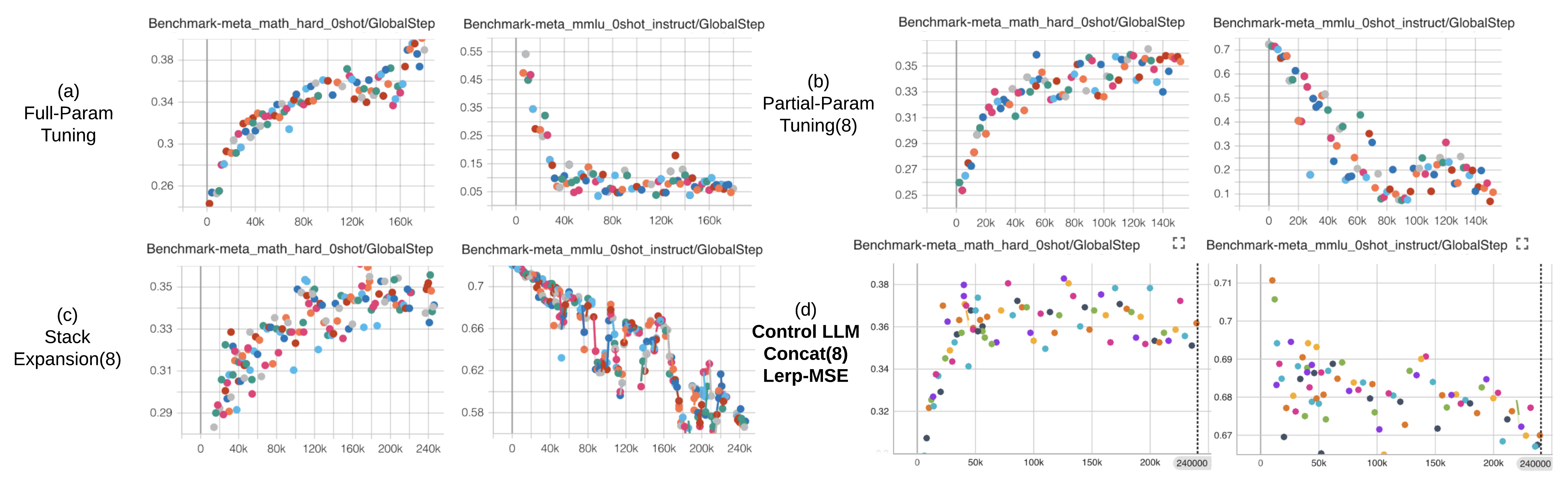} 
    \vspace{-5pt}
    \caption{\textbf{[OpenMath + Llama-3.1-8B-Instruct]} Benchmark comparison of training methods. \textbf{(a)} Full Parameter Tuning. \textbf{(b)} Partial Parameter Tuning: Freeze all except 1 of every 4 transformer layers (8 total). \textbf{(c)} Stack Expansion: Add 8 transformer layers (1 per 4) while freezing originals. \textbf{(d)} Concat-Lerp Expansion: Add 8 transformer layers connected via interpolator with MSE divergence loss.}

    \label{fig:control_llm_training_plot_comparison_math}
\end{figure*}

\begin{figure*}[ht]
    \centering
    \includegraphics[width=6.51in]{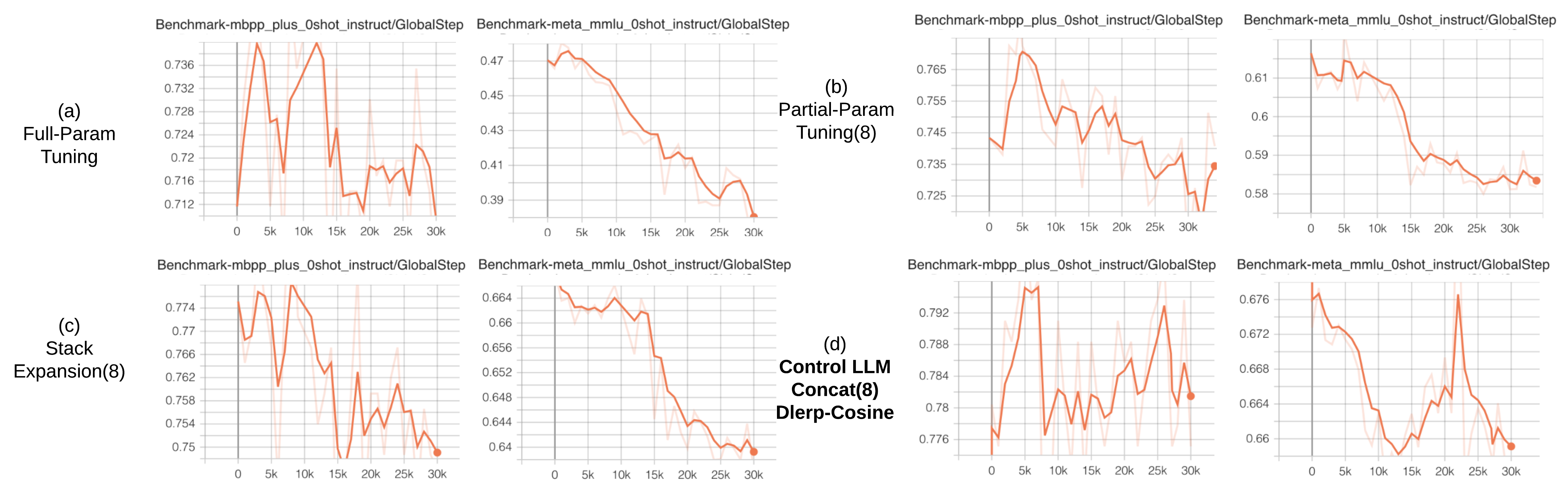} 
    \vspace{-5pt}
    \caption{\textbf{[OpenCoder SFT Phase2 + Llama-3.1-8B-Instruct]} Benchmark comparison of training methods. \textbf{(a)} Full Parameter Tuning. \textbf{(b)} Partial Parameter Tuning: Freeze all except 1 of every 4 transformer layers (8 total). \textbf{(c)} Stack Expansion: Add 8 transformer layers (1 per 4) while freezing originals. \textbf{(d)} Concat-Lerp Expansion: Add 8 transformer layers connected via interpolator with MSE divergence loss.}

    \label{fig:control_llm_training_plot_comparison_code}
\end{figure*}


Benchmark results in Tables~\ref{tab:csft_results_math}, \ref{tab:csft_results_coding}, \ref{tab:cpt_results_multilingual}, and \ref{tab:ablation_study} demonstrate that Control LLM effectively mitigates catastrophic forgetting (CF) while enabling the learning of new skills across mathematical, coding, and multilingual tasks using task-specific data. These results reflect the performance of the best-performing checkpoints. To provide a more granular view, we plot benchmark results at every 1K training steps, highlighting the \textbf{learn more, forget less} principle and comparing different fine-tuning methods.


\subsection{Mathematical Benchmarks}

Figure~\ref{fig:control_llm_training_plot_comparison_math} illustrates the exact match accuracy for MathHard-0shot (representing new skills) and MMLU-0shot (representing original capabilities) during the fine-tuning of Llama-3.1-8B-Instruct~\cite{meta2024llama31instruct} on the OpenMath2 dataset~\cite{nvidia2024openmathinstruct2}. Key observations include:

\begin{itemize}[leftmargin=*]
    \item \textbf{Full Parameter Tuning}: Rapidly improves MathHard accuracy from 23.7\% to 38\% within 170K steps. However, it causes severe degradation in Original Capabilities (MMLU drops below 15\% from 72.4\%) within 40K steps. Other tasks like GSM8K and ARC exhibit similar trends.
    \item \textbf{Partial Parameter Tuning}: Slows degradation of Original Capabilities, maintaining MMLU accuracy above 20\% for 80K steps, but fails to sustain long-term performance.
    \item \textbf{Stack Expansion}: Further improves retention, maintaining MMLU accuracy above 50\% for 240K steps while converging on MathHard accuracy.
    \item \textbf{Control LLM (Concat-Lerp with MSE Loss)}: Maintains the best balance, preserving MMLU accuracy above 67\% even after 240K steps, while achieving 38\% MathHard accuracy.
\end{itemize}


\subsection{Coding Benchmarks}

Figure~\ref{fig:control_llm_training_plot_comparison_code} presents results from fine-tuning on the smaller-scale OpenCoder-SFT-Phase2 dataset (245.4M tokens, compared to OpenMath2’s 5.1B tokens). Key trends include:

\begin{itemize}[leftmargin=*]
    \item \textbf{Full Parameter Tuning}: Struggles to converge to optimal coding performance (MBPP-Plus improves only slightly from 70.4\% to 73.6\%) while degrading Original Capabilities (MMLU drops below 40\%) within 30K steps. Similar trends are observed for HumanEval.
    \item \textbf{Partial Parameter Tuning}: Achieves moderate improvement, with MBPP-Plus reaching 76.5\% accuracy and MMLU accuracy staying above 50\% for 30K steps.
    \item \textbf{Stack Expansion}: Converges to 77.4\% MBPP-Plus accuracy while preserving MMLU accuracy above 64\%.
    \item \textbf{Control LLM}: Achieves the best performance, converging to 79.2\% MBPP-Plus accuracy while maintaining MMLU accuracy above 66\%.
\end{itemize}

\subsection{Notes on Coding Results}

The plotting experiment in Figure~\ref{fig:control_llm_training_plot_comparison_code} exclusively uses OpenCoder-SFT-Phase2 data. In contrast, the results reported in Table~\ref{tab:csft_results_coding} use both OpenCoder-SFT-Phase1 and Phase2 datasets in a 2-phase SFT approach, achieving the optimal 80\% MBPP-Plus performance.

\section{Hidden State Alignment}
\label{sec:probing}


\subsection{Hidden State Probing}
\label{sec:probing_data}


To investigate the relationship between hidden-state alignment and model performance, we crafted probing sentences such as "king is to queen" and passed them through a CSFT-trained model. Table~\ref{tab:probe_data} provides examples of the probe data used. Hidden states of the final tokens were projected layer-wise, labeled as [Expanded] for the expanded transformer block and [Pre-trained] for the original pre-trained block. Using PCA, we visualized these hidden states in a 3D embedding space. Projections of nearest neighbors for specific pairs such as King-Queen, Swim-Swam, or Paris-France reveal that semantically related representations form tight clusters due to extensive pre-training on large-scale datasets.


\begin{table}[ht]
  \centering
  \caption{Categorized Probe Data with Compiled Sentences}
  \label{tab:probe_data}
  {\small
  \begin{tabularx}{\linewidth}{>{\bfseries}l p{0.6\linewidth} p{0.3\linewidth}}
    \hline
    \textbf{Category} & \textbf{Analogous Word Pairs} & \textbf{Sentences(Examples)} \\
    \hline
    Gender Pairs        & king:queen, man:woman, actor:actress, waiter:waitress, uncle:aunt, nephew:niece & king is to queen \\
    Verb Tenses         & swim:swam, fly:flew, see:saw, go:went, walking:walked, swimming:swam & swim is to swam \\
    Capital-Country     & Paris:France, Tokyo:Japan, Brasilia:Brazil, Ottawa:Canada         & Paris is to France \\
    Country-Currency    & Russia:ruble, Japan:yen, United States:dollar, United Kingdom:pound & Russia is to ruble \\
    Singular-Plural     & dog:dogs, cat:cats, mouse:mice, goose:geese, child:children, person:people & mouse is to mice \\
    Comparative Forms   & good:better, cold:colder                                          & good is to better \\
    Entity-Product      & Apple:iPhone, Microsoft:Windows                                   & Apple is to iPhone \\
    Language-Nationality& Spain:Spanish, Italy:Italian                                      & Spain is to Spanish \\
    Family Relations    & uncle:aunt, nephew:niece                                         & uncle is to aunt \\
    Other Relations     & tree:forest, building:city                                       & tree is to forest \\
    \hline
  \end{tabularx}
  }
\end{table}

\begin{figure*}[htbp] 
  \centering
  \begin{subfigure}[t]{0.48\textwidth}
    \centering
    \includegraphics[height=5cm, keepaspectratio]{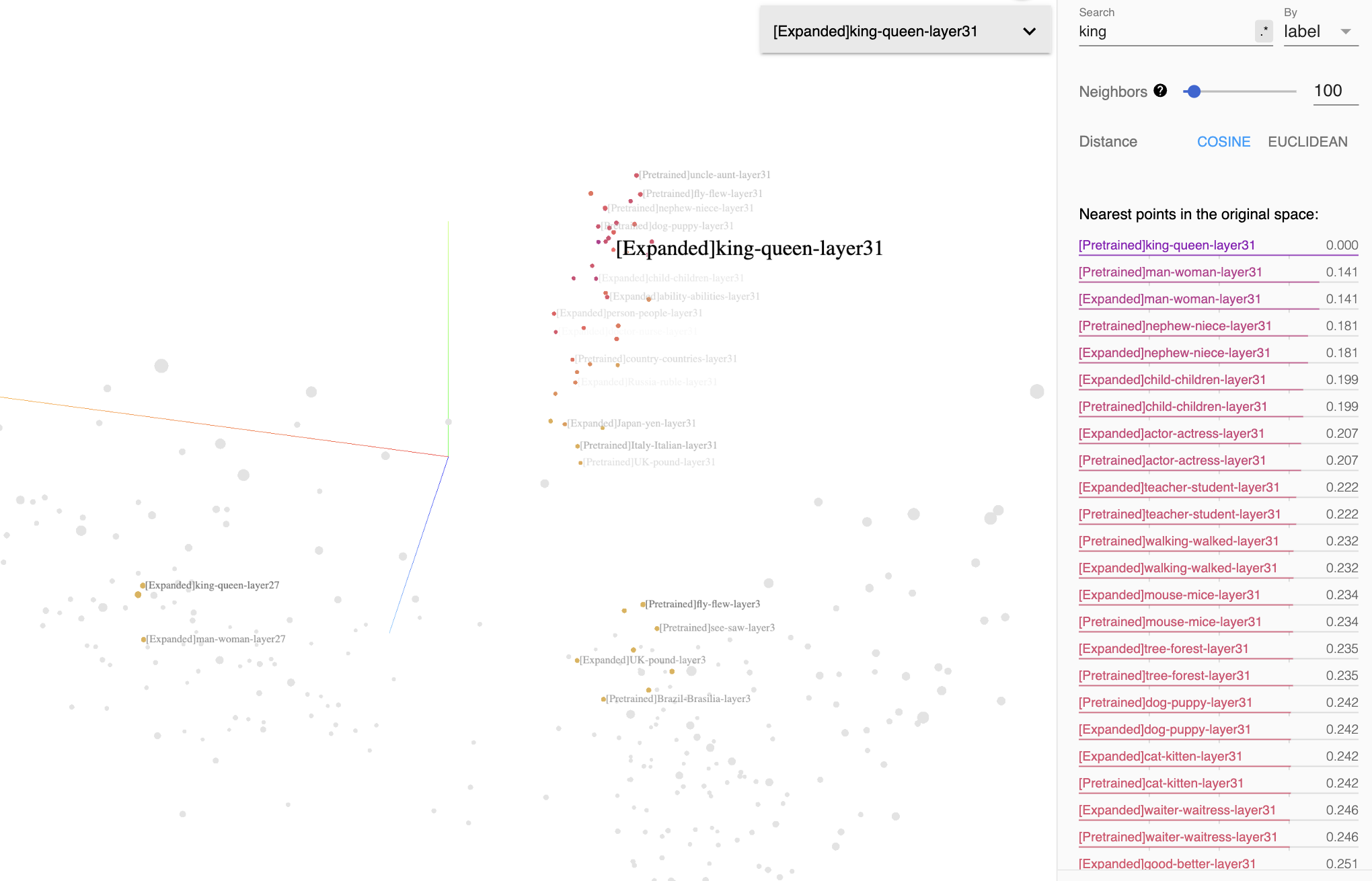}
    \caption{Pre-trained Model - Math Hard 0.237 - MMLU 0.724}
    \label{fig:alignment_pretrain}
  \end{subfigure}
  \hfill
  \begin{subfigure}[t]{0.48\textwidth}
    \centering
    \includegraphics[height=5cm, keepaspectratio]{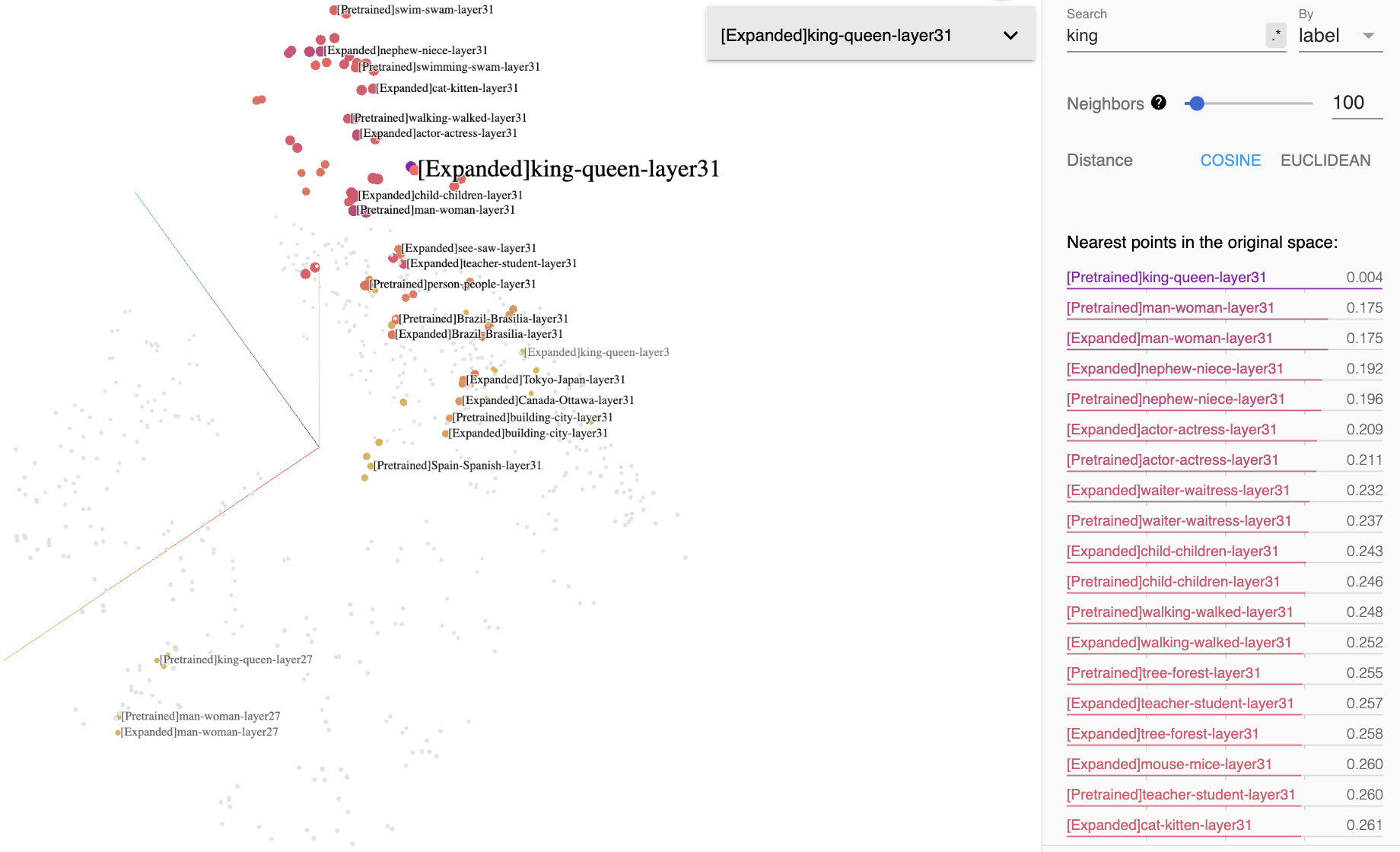}
    \caption{Lerp with MSE - Math Hard 0.360 - MMLU 0.716}
    \label{fig:alignment_best}
  \end{subfigure}
  
  \begin{subfigure}[t]{0.48\textwidth}
    \centering
    \includegraphics[height=5cm, keepaspectratio]{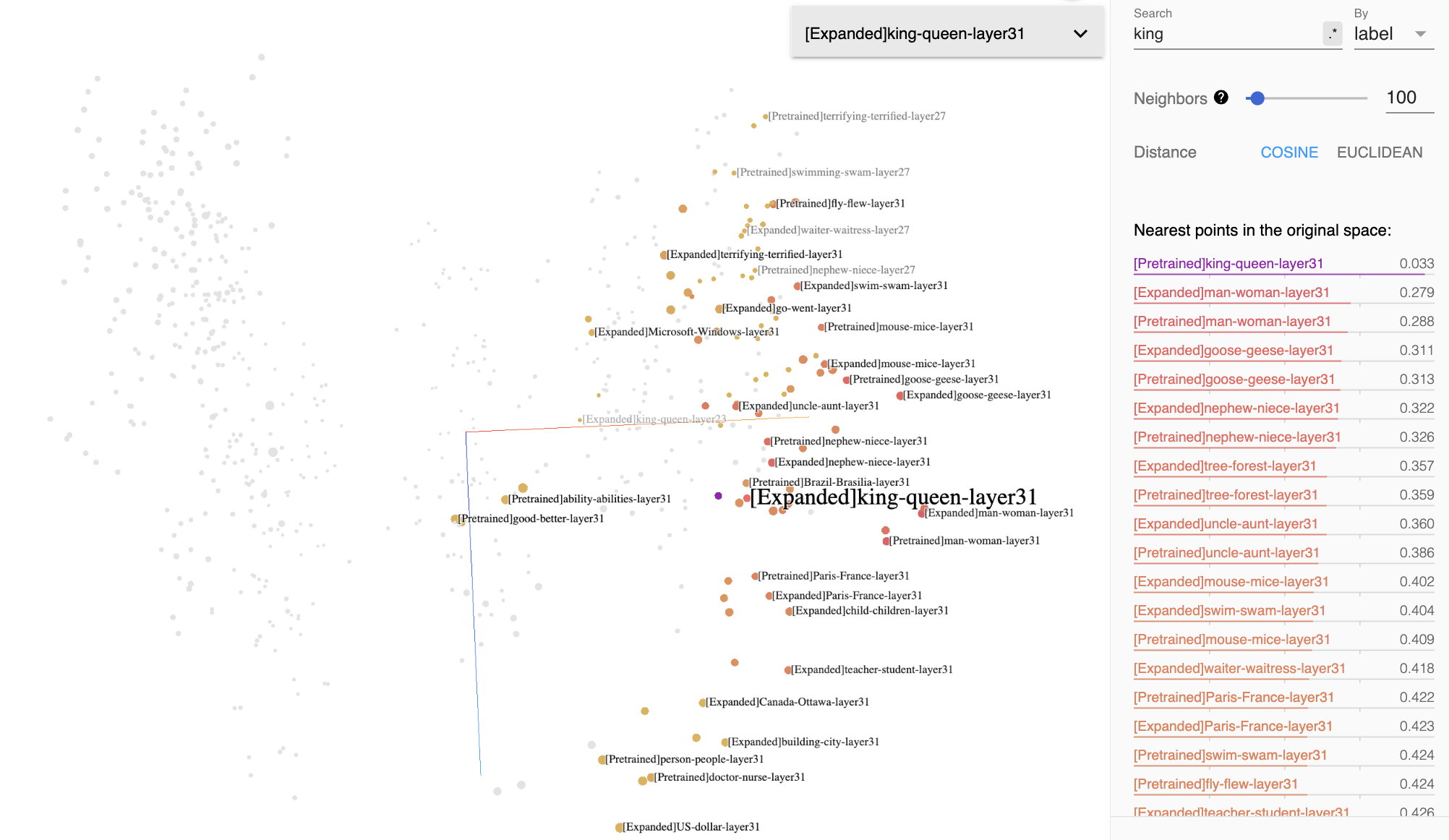}
    \caption{Dlerp without Divergence Loss - Math Hard 0.357 - MMLU 0.66}
    \label{fig:alignment_good}
  \end{subfigure}
  \hfill
  \begin{subfigure}[t]{0.48\textwidth}
    \centering
    \includegraphics[height=5cm, keepaspectratio]{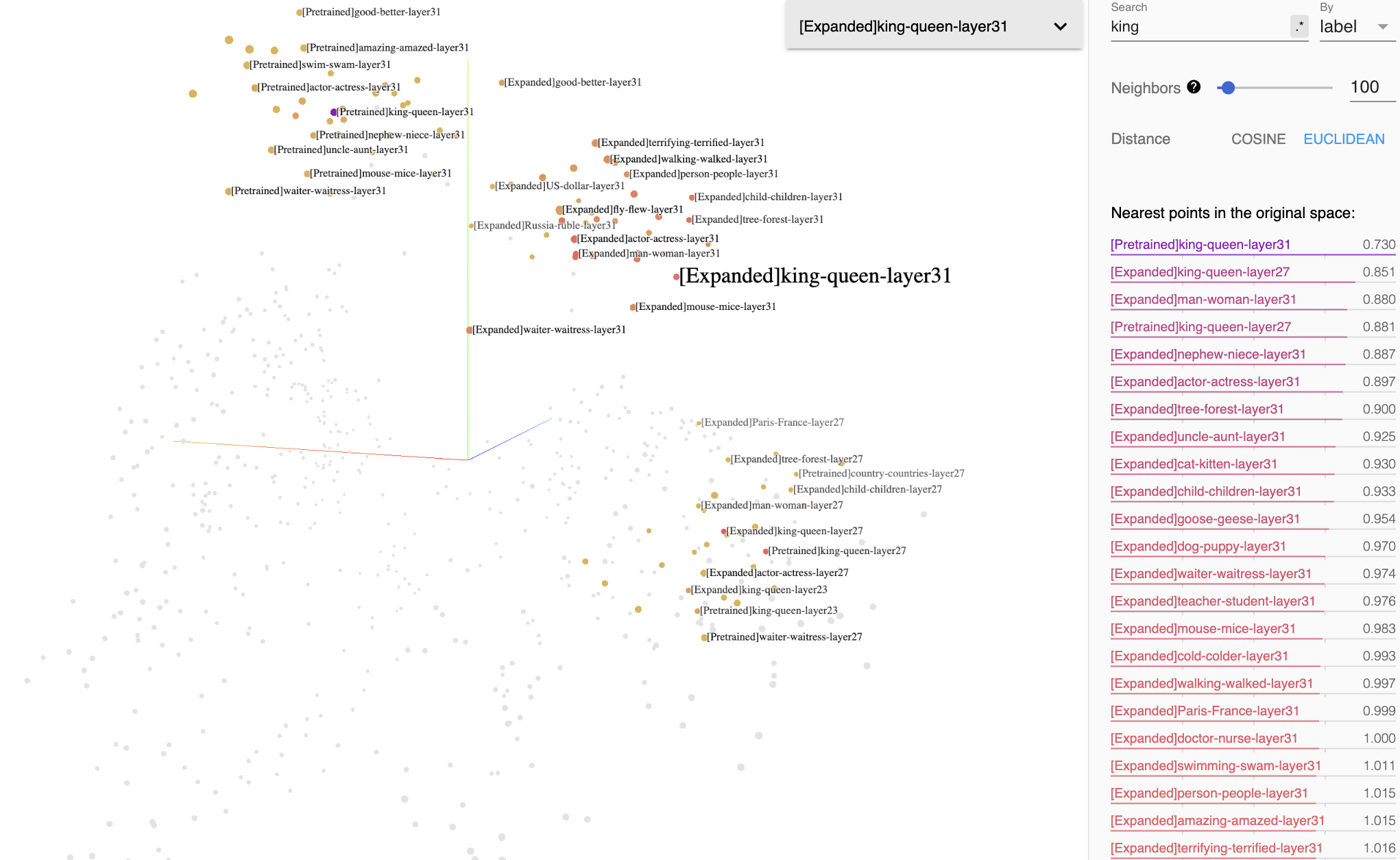}
    \caption{Lerp with Cosine Alignment - Math Hard 0.362 - MMLU 0.54}
    \label{fig:alignment_cosine}
  \end{subfigure}

  \begin{subfigure}[t]{0.48\textwidth}
    \centering
    \includegraphics[height=5cm, keepaspectratio]{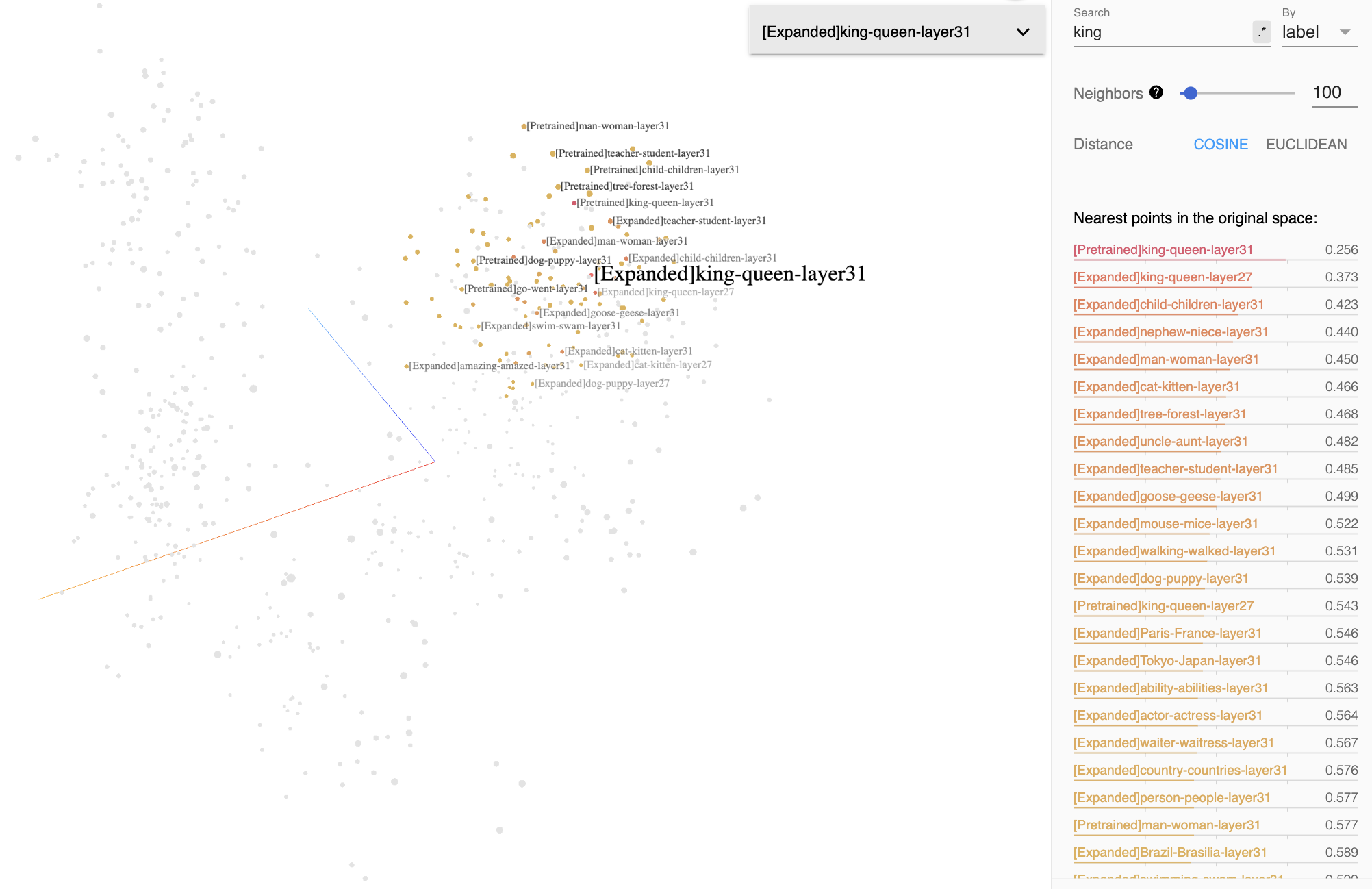}
    \caption{Lerp without Divergence Loss - Math Hard 0.359 - MMLU 0.41}
    \label{fig:alignment_no_div}
  \end{subfigure}
  \hfill
  \begin{subfigure}[t]{0.48\textwidth}
    \centering
    \includegraphics[height=5cm, keepaspectratio]{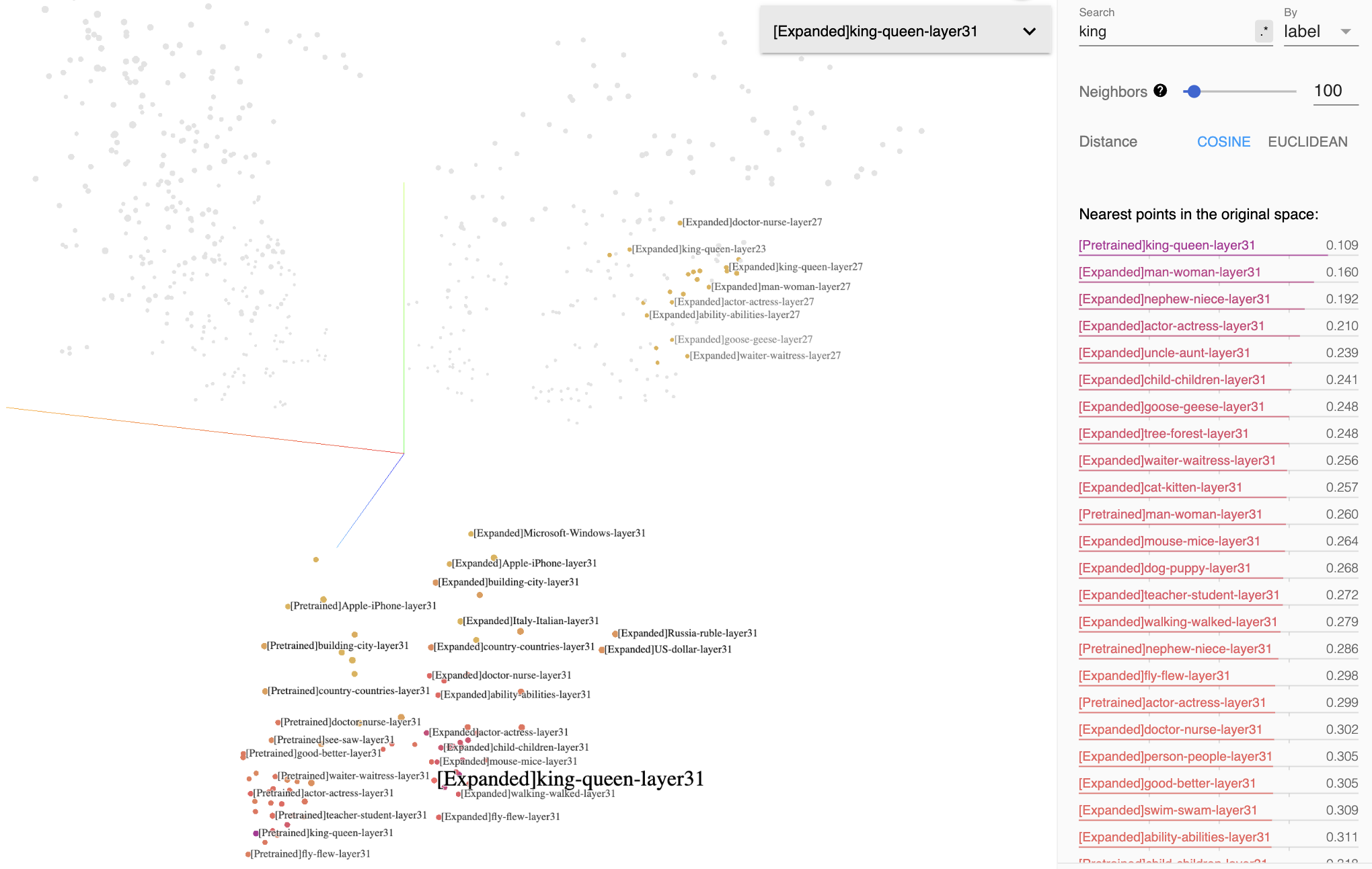}
    \caption{Full Parameter Training - Math Hard 0.368 - MMLU 0.07}
    \label{fig:alignment_worst}
  \end{subfigure}
  

    \caption{\textbf{[OpenMath2 + Llama-3.1-8B-Instruct]} Comparison of various alignment strategies and their impact on model performance. 
    \textbf{(a)} \textbf{Pre-trained Model:} Hidden states from [Pretrained] and [Expanded] blocks are identical for all layers. 1) \textit{Semantic Stability:} Analogous sentences exhibit high similarity in both cosine and Euclidean distances. 2) \textit{Layer-wise Clustering:} Distinct clusters are formed for each layer. 
    \textbf{(b)} \textbf{Control LLM (MSE Divergence Loss):} Maintains both semantic stability and layer-wise clustering, exhibiting strong catastrophic forgetting (CF) mitigation. 
    \textbf{(c)} \textbf{Dlerp without Divergence Loss:} Shows distances approximately 2x larger than (b), reducing alignment quality. 
    \textbf{(d)} \textbf{Lerp with Cosine Divergence Loss:} Preserves layer-wise clustering but weakens semantic stability. Nearest neighbors for [Pretrained] drift in Euclidean distance, while cosine similarity is maintained. 
    \textbf{(e)} \textbf{Lerp without Divergence Loss:} Semantic stability degrades further between [Pretrained] and [Expanded] blocks, particularly in cosine similarity. 
    \textbf{(f)} \textbf{Full Parameter Tuning:} Semantic drift is observed between [Expanded] and [Pretrained], with no interpolation applied.}

  \label{fig:alignment_group}
\end{figure*}

\subsection{Hidden State Probing Result}
\label{sec:probing_result}


Figure~\ref{fig:alignment_comparison} compares Best and Worst Alignment by examining the output hidden states across transformer layers for various categories of probe data. This highlights the correlation between catastrophic forgetting (CF) and hidden-state alignment, as reflected in the degradation of original capabilities. The following observations are derived from fine-tuning experiments:

\subsection{Correlation Between Alignment and Catastrophic Forgetting}
\label{sec:alignment_results}

Figure~\ref{fig:alignment_group} presents the relationship between alignment quality and model performance across five fine-tuning experiments. Each experiment involved fine-tuning Llama-3.1-8B-Instruct on the OpenMath2 dataset. We analyze the final checkpoints for each model, which have converged on the Math-Hard benchmark. Key findings are:

\begin{itemize}[leftmargin=*]
    \item \textbf{(a) Pre-trained Model:} The pre-trained checkpoint serves as the baseline, where [Pretrained] and [Expanded] hidden states are identical across layers. Observations include:
    \begin{enumerate}[leftmargin=*]
        \item \textbf{Semantic Stability:} Analogous sentences exhibit high similarity in both cosine and Euclidean distances.
        \item \textbf{Layer-wise Clustering:} Hidden states form distinct clusters for each layer.
    \end{enumerate}
    The model retains original capabilities with no CF.

    \item \textbf{(b) Control LLM (Lerp + MSE Divergence Loss):} Both Semantic Stability and Layer-wise Clustering are preserved. Distances between [Pretrained] and [Expanded] are minimal, yielding strong CF mitigation. MMLU drops slightly from 0.724 to 0.716, demonstrating minimal forgetting.

    \item \textbf{(c) Control LLM (Dlerp without Divergence Loss):} Distances between [Pretrained] and [Expanded] are approximately 2x larger than (b). However, the [Pretrained] and [Expanded] representations of the same sentence in the same layer remain closely aligned.

    \item \textbf{(d) Control LLM (Lerp + Cosine Divergence Loss):} Layer-wise clustering is preserved, but Semantic Stability is weaker. Nearest neighbors of [Pretrained] drift slightly in Euclidean distance, although cosine similarity remains high. MMLU drops significantly to 0.54.

    \item \textbf{(e) Control LLM (Lerp without Divergence Loss):} Semantic Stability deteriorates further, with cosine similarity between [Pretrained] and [Expanded] dropping. [Pretrained] tags no longer appear in the top 20 nearest neighbors. MMLU falls to 0.41 but remains better aligned than (f) due to the interpolation strategy.

    \item \textbf{(f) Full Parameter Tuning:} Without interpolation, [Pretrained] and [Expanded] representations exhibit the worst Semantic Stability. Catastrophic Forgetting is most pronounced, with MMLU dropping to 0.07.
\end{itemize}

\paragraph{Summary of Findings.} These results demonstrate that alignment quality directly impacts the balance between learning new tasks and preserving prior knowledge. Control LLM configurations with interpolation mechanisms, particularly Lerp with MSE divergence loss, achieve the best trade-off, minimizing CF while ensuring strong performance on new tasks.

\section{Evaluation Details}
\label{sec:eval_details}

We evaluate Control LLM across multiple benchmarks. Table~\ref{tab:eval_datasets} lists the datasets, abbreviations, reference URLs to the data sources and the prompt template.

\begin{table}[!ht]
\centering
\caption{Evaluation datasets, abbreviations, data source, and prompt templates. $*$ - input\_final\_prompts column}
\label{tab:eval_datasets}
\begin{tabular}{@{}lp{5cm}p{3cm}p{3cm}@{}}
\toprule
\textbf{Task}         & \textbf{Dataset (Abbreviation)}       & \textbf{Data Source}           & \textbf{Prompt Template}                                                    \\ \midrule
\textbf{Math}         & Math-0shot (Math)                    & \href{#ref1}{[1]}            & \href{#ref15}{[15]}*                                       \\
                      & Math-Hard-0shot (MathH)              & \href{#ref2}{[2]}            & \href{#ref16}{[16]}*                                             \\
                      & GSM8K (G8K)                          & \href{#ref3}{[3]}            & \href{#ref17}{[17]}*                                \\ \midrule
\textbf{Coding}       & MBPP-Sanitize-0shot (MBPPS)          & \href{#ref4}{[4]}            & \href{#ref18}{[18]}*                              \\
                      & MBPP-Plus-0shot (MBPP+)              & \href{#ref5}{[5]}            & \href{#ref19}{[19]}*                          \\
                      & HumanEval-Greedy (HE)                & \href{#ref6}{[6]}            & \href{#ref20}{[20]}*                                \\
                      & HumanEval-Plus-Greedy (HE+)          & \href{#ref7}{[7]}            & \href{#ref21}{[21]}*                           \\ \midrule
\textbf{Chinese}      & C-Eval-0shot (CEval)                 & \href{#ref8}{[8]}            & \href{#ref22}{[22]}*                        \\
                      & C-Eval-0shot-CoT (CEvalC)            & \href{#ref8}{[8]}            & \ref{eval:zh_cot_prompt}                              \\
                      & CMMLU-0shot (CMMLU)                  & \href{#ref9}{[9]}            & \href{#ref23}{[23]}*                              \\
                      & CMMLU-0shot-CoT (CMMLUC)             & \href{#ref9}{[9]}            & \ref{eval:zh_cot_prompt}                                      \\ \midrule
\textbf{Original Capabilities-CSFT} & ARC\_Challenge-0shot (ARC) & \href{#ref10}{[10]}          & \href{#ref10}{[10]}*                              \\
                      & GPQA-0shot (GPQA)                   & \href{#ref11}{[11]}          & \href{#ref11}{[11]}*                                        \\
                      & MMLU-0shot (MMLU)                   & \href{#ref12}{[12]}          & \href{#ref12}{[12]}*                                    \\
                      & MMLU\_Pro-5shot (MMLUP)             & \href{#ref13}{[13]}          & \href{#ref13}{[13]}*                             \\ \midrule
\textbf{Original Capabilities-CPT}  & BBH-3shot (BBH)           & \href{#ref14}{[14]}          & \href{#ref14}{[14]}*                                 \\
                      & MMLU-5shot (MMLU)                   & \href{#ref12}{[12]}          & \href{#ref12}{[12]}*                                 \\
                      & MMLU\_Pro-5shot (MMLUP)             & \href{#ref13}{[13]}          & \href{#ref13}{[13]}*                      \\ \bottomrule
\end{tabular}
\end{table}

\makeatletter
\renewcommand{\@makefntext}[1]{%
  \noindent\raggedright\fontsize{5.9}{5}\selectfont\@thefnmark.\, #1%
}
\makeatother

\footnotetext[1]{\url{https://huggingface.co/datasets/lighteval/MATH}}
\footnotetext[2]{\url{https://huggingface.co/datasets/lighteval/MATH-Hard}}
\footnotetext[3]{\url{https://huggingface.co/datasets/openai/gsm8k}}
\footnotetext[4]{\url{https://huggingface.co/datasets/google-research-datasets/mbpp/viewer/sanitized}}
\footnotetext[5]{\url{https://huggingface.co/datasets/evalplus/mbppplus}}
\footnotetext[6]{\url{https://huggingface.co/datasets/openai/openai_humaneval}}
\footnotetext[7]{\url{https://huggingface.co/datasets/evalplus/humanevalplus}}
\footnotetext[8]{\url{https://huggingface.co/datasets/ceval/ceval-exam}}
\footnotetext[9]{\url{https://huggingface.co/datasets/haonan-li/cmmlu}}
\footnotetext[10]{\url{https://huggingface.co/datasets/meta-llama/Llama-3.1-8B-Instruct-evals/viewer/Llama-3.1-8B-Instruct-evals__arc_challenge__details}}
\footnotetext[11]{\url{https://huggingface.co/datasets/meta-llama/Llama-3.1-8B-Instruct-evals/viewer/Llama-3.1-8B-Instruct-evals__gpqa__details}}
\footnotetext[12]{\url{https://huggingface.co/datasets/meta-llama/Llama-3.1-8B-Instruct-evals/viewer/Llama-3.1-8B-Instruct-evals__mmlu__0_shot__cot__details}}
\footnotetext[13]{\url{https://huggingface.co/datasets/meta-llama/Llama-3.1-8B-Instruct-evals/viewer/Llama-3.1-8B-Instruct-evals__mmlu_pro__details}}
\footnotetext[14]{\url{https://huggingface.co/datasets/meta-llama/Llama-3.1-8B-evals/viewer/Llama-3.1-8B-evals__bbh__details}}
\footnotetext[15]{\url{https://huggingface.co/datasets/meta-llama/Llama-3.1-8B-Instruct-evals/viewer/Llama-3.1-8B-Instruct-evals__math__details}}
\footnotetext[16]{\url{https://huggingface.co/datasets/meta-llama/Llama-3.1-8B-Instruct-evals/viewer/Llama-3.1-8B-Instruct-evals__math_hard__details}}
\footnotetext[17]{\url{https://huggingface.co/datasets/meta-llama/Llama-3.1-8B-Instruct-evals/viewer/Llama-3.1-8B-Instruct-evals__gsm8k__details}}
\footnotetext[18]{\url{https://huggingface.co/datasets/meta-llama/Llama-3.1-8B-Instruct-evals/viewer/Llama-3.1-8B-Instruct-evals__mbpp__details}}
\footnotetext[19]{\url{https://huggingface.co/datasets/meta-llama/Llama-3.1-8B-Instruct-evals/viewer/Llama-3.1-8B-Instruct-evals__mbpp_plus__details}}
\footnotetext[20]{\url{https://huggingface.co/datasets/meta-llama/Llama-3.1-8B-Instruct-evals/viewer/Llama-3.1-8B-Instruct-evals__human_eval__details}}
\footnotetext[21]{\url{https://huggingface.co/datasets/meta-llama/Llama-3.1-8B-Instruct-evals/viewer/Llama-3.1-8B-Instruct-evals__human_eval_plus__details}}
\footnotetext[22]{\url{https://github.com/EleutherAI/lm-evaluation-harness/blob/main/lm_eval/tasks/ceval/_default_ceval_yaml}}
\footnotetext[23]{\url{https://github.com/EleutherAI/lm-evaluation-harness/blob/main/lm_eval/tasks/cmmlu/_default_template_yaml}}

\makeatletter
\renewcommand{\@makefntext}[1]{\noindent\@thefnmark.\, #1}
\makeatother

\subsection{Chain-of-Thought Evaluation for Chinese Benchmarks}
\label{eval:zh_cot_prompt}
For CEvalC and CMMLUC, chain-of-thought instruction following is specifically assessed in Chinese. These benchmarks utilize the same datasets as C-Eval and CMMLU, with the prompt in Table~\ref{tab:chinese_prompt}.

\begin{CJK}{UTF8}{gbsn} 
\begin{table}[!ht]
\centering
\caption{Chain-of-Thought Prompt for Chinese Benchmarks - CEvalC and CMMLUC}
\label{tab:chinese_prompt}
\begin{tabular}{|p{0.9\textwidth}|}
\hline
在以下问题和四个候选答案（A、B、C 和 D）中，选择最佳答案。
问题：... \\
A. ... \\
B. ... \\
C. ... \\
D. ... \\
对于简单的问题：尽量简洁地解释, 并提供答案。对于复杂的问题：使用以下逐步格式：\\
\textbf{步骤 1:} [简明描述] [简要解释] \\
\textbf{步骤 2:} [简明描述] [简要解释] \\
无论采用哪种方法，始终以以下内容结束： 最佳答案是 [答案字母]。[答题结束] 其中 [答案字母] 是 A、B、C 或 D 中的一个。
让我们一步一步思考。
\end{tabular}
\end{table}
\end{CJK}

This prompt ensures structured reasoning for complex problems and concise answers for simpler ones, promoting consistent evaluation of chain-of-thought capabilities in Chinese tasks.


\section{Ablation Study}
\label{sec:ablation_study}

\begin{figure*}[ht]
    \centering
    \includegraphics[width=6.51in]{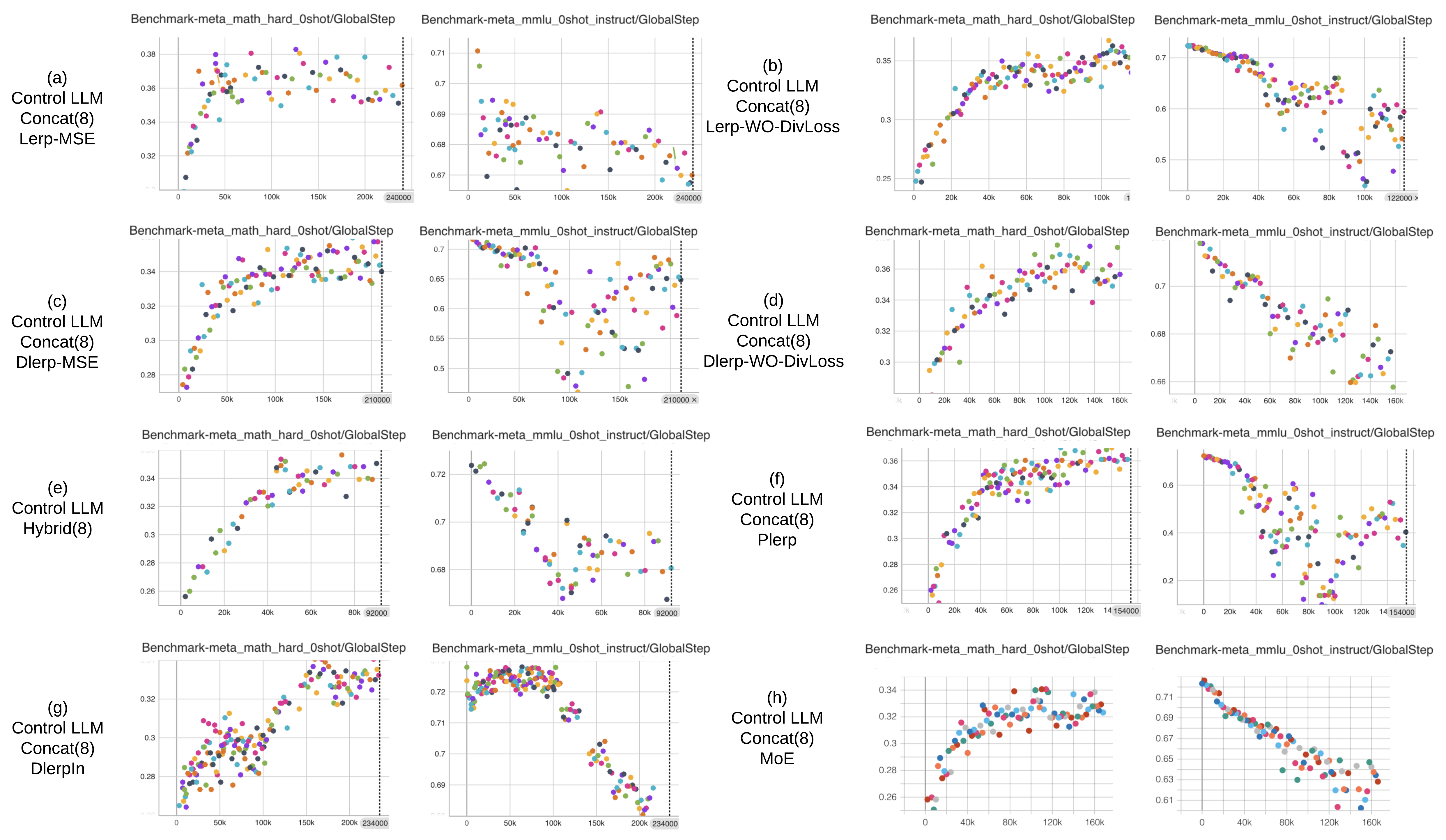} 
    \vspace{-5pt}
    \caption{\textbf{[OpenMath2 + Llama-3.1-8B-Instruct]} Comparison of benchmark results of different ablation study settings every 1K steps during training. \textbf{(a)} Lerp Interporation Strategy with MSE loss. \textbf{(b)} Lerp Interporation Strategy without Divergence Loss. \textbf{(c)} Dlerp Interporation Strategy with MSE loss. \textbf{(d)} Dlerp Interporation Strategy without Divergence Loss. \textbf{(e)} Hybrid Expansion Strategy. \textbf{(e)} Plerp Interporation Strategy. \textbf{(f)} DlerpIn Interporation Strategy. \textbf{(e)} MoE gating. }
    \label{fig:control_llm_training_plot_ablation_math}
\end{figure*}

\begin{figure*}[ht]
    \centering
    \includegraphics[width=6.51in]{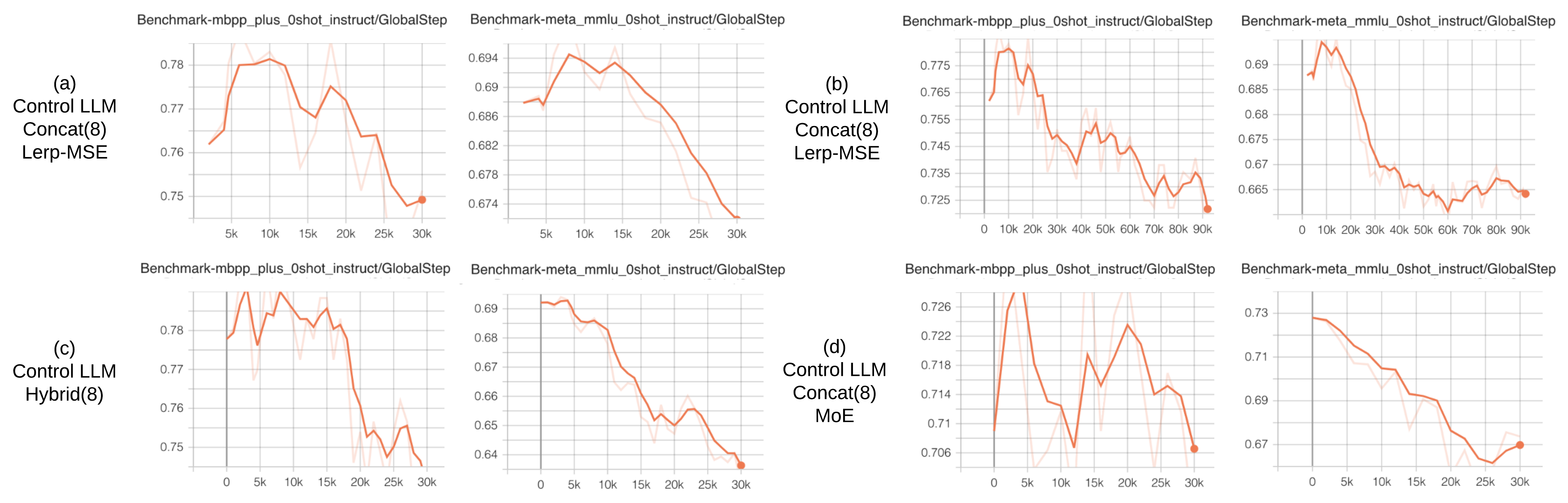} 
    \vspace{-5pt}
    \caption{\textbf{[OpenCoder + Llama-3.1-8B-Instruct]} Comparison of benchmark results of different ablation study settings during training. \textbf{(a)} Lerp Interporation Strategy with MSE Loss. \textbf{(b)} Lerp Interporation Strategy with MSE Loss. Trained 3X longer than (a) \textbf{(c)} Hybrid Expansion Strategy. \textbf{(d)} MoE gating.}
    \label{fig:control_llm_training_plot_ablation_code}
\end{figure*}

This section examines the impact of various settings in mitigating Catastrophic Forgetting (CF) and learning new tasks through an ablation study. Figures~\ref{fig:control_llm_training_plot_ablation_math} and \ref{fig:control_llm_training_plot_ablation_code} present benchmark results recorded at every 1K steps during training.

\subsection{Ablation Study on Mathematical Tasks}
\label{sec:ablation_math}

Figure~\ref{fig:control_llm_training_plot_ablation_math} shows results from tuning Llama-3.1-8B-Instruct on the OpenMath2 dataset. Key findings for different configurations include:

\begin{itemize}[leftmargin=*]
    \item \textbf{(a) Lerp + MSE (Concat Expansion):} Using 8 transformer blocks (1 every 4 layers) and MSE divergence loss, this configuration maintains MMLU above 67\%, demonstrating strong CF mitigation. Compared to \textbf{(b)} Lerp without divergence loss, MMLU drops below 50\%, highlighting the importance of MSE in mitigating CF. 
    \item \textbf{(c) Dlerp + MSE:} This setting achieves comparable CF mitigation, keeping MMLU above 65\%. Although performance fluctuates due to dynamic interpolation, checkpoints can be easily selected to optimize both new tasks (e.g., Math-Hard) and Original Capabilities (e.g., MMLU).
    \item \textbf{(d) Dlerp without Divergence Loss:} Without MSE, performance is less stable, and MMLU drops below the levels observed in \textbf{(c)}.
    \item \textbf{(e) Hybrid Expansion Strategy:} Similar to \textbf{(a)} and \textbf{(c)}, this strategy delivers strong results for both new tasks and CF mitigation.
    \item \textbf{(f) Plerp vs. (g) DlerpIn:} Plerp excels at learning new tasks but performs poorly in CF mitigation. Conversely, DlerpIn outperforms Plerp in preserving Original Capabilities, maintaining MMLU above 68\% after 234K steps. 
    \item \textbf{(h) MoE Gating:} Although effective in CF mitigation, MoE's MMLU drops to ~63\% after 160K steps, underperforming DlerpIn.
\end{itemize}


\subsection{Ablation Study on Coding Tasks}
\label{sec:ablation_code}

Figure~\ref{fig:control_llm_training_plot_ablation_code} presents the results of tuning Llama-3.1-8B-Instruct on the OpenCoder-SFT-Phase2 dataset. This dataset is smaller in scale (245.4M tokens) compared to OpenMath2 (5.1B tokens). Key observations include:

\begin{itemize}[leftmargin=*]
    \item \textbf{(a) Lerp + MSE:} Achieves performance comparable to Dlerp-Cosine (Figure~\ref{fig:control_llm_training_plot_comparison_code}-(d)). MMLU remains above 66\% even after 90K steps, showing strong CF mitigation.
    \item \textbf{(b) Extended Training:} Training 3x longer demonstrates that MMLU stability persists, reinforcing the robustness of the Lerp + MSE configuration.
    \item \textbf{(c) Hybrid Expansion:} Delivers solid performance in both learning new tasks and mitigating CF.
    \item \textbf{(d) MoE Gating:} While effective for CF mitigation, MoE underperforms in learning new tasks compared to other proposed methods.
\end{itemize}


\subsection{Summary of Findings}

This ablation study underscores the importance of Control LLM's architecture, including its expansion strategies, interpolation techniques, and divergence loss. Key insights are:
\begin{itemize}[leftmargin=*]
    \item \textbf{Expansion Strategies:} Concat and Hybrid strategies outperform Stack, as they enable interpolation and make divergence loss effective. Stack is still superior to full parameter tuning, which is widely used.
    \item \textbf{Interpolation Methods:} Lerp and Dlerp outperform MoE, a method commonly adopted by open-source models. Even DlerpIn, a "soft" version of MoE, proves more effective in mitigating CF.
    \item \textbf{Divergence Loss:} The inclusion of divergence loss, particularly MSE, significantly enhances CF mitigation while maintaining stability during training.
\end{itemize}

\section{Acknowledgments}

We thank the teams of Llama3\citep{dubey2024llama}, Llama3-SynE\citep{chen2024towards}, OpenMathInstruct-2\citep{toshniwal2024openmathinstruct}, OpenCoder\citep{huang2024opencoder}, llama-recipes\citep{meta2024llamarecipes} and lm-eval-harvness\citep{eleutherai2024evaluation} for pre-trained model, datasets, tools and resources that contributed to this research.

\end{document}